\documentclass{article}

     \PassOptionsToPackage{numbers, compress}{natbib}
\usepackage[preprint]{neurips_2022}

\usepackage[utf8]{inputenc} % allow utf-8 input
\usepackage[T1]{fontenc}    % use 8-bit T1 fonts
\usepackage[hidelinks]{hyperref}       % hyperlinks
\usepackage{url}            % simple URL typesetting
\usepackage{booktabs}       % professional-quality tables
\usepackage{amsfonts}       % blackboard math symbols
\usepackage{nicefrac}       % compact symbols for 1/2, etc.
\usepackage{microtype}      % microtypography
\usepackage{xcolor}         % colors
\usepackage{wrapfig}
\usepackage{tabularx}
\usepackage{bbding}

\usepackage{times}
\usepackage{latexsym}

\usepackage[utf8]{inputenc} % allow utf-8 input
\usepackage[T1]{fontenc}    % use 8-bit T1 fonts
\usepackage{hyperref}       % hyperlinks
\hypersetup{
    colorlinks=true,
    linkcolor=black,
    urlcolor=blue
    }
\usepackage{url}            % simple URL typesetting
\usepackage{booktabs}       % professional-quality tables
\usepackage{amsfonts}       % blackboard math symbols
\usepackage{nicefrac}       % compact symbols for 1/2, etc.
\usepackage{microtype}      % microtypography
\usepackage{amsmath,amssymb}
\usepackage{epsfig,graphicx,subfigure,caption}
\usepackage{algpseudocode}
\usepackage{multirow,booktabs, hhline}
\usepackage[ruled,noend]{algorithm2e}
\usepackage{amsmath, bm}
\usepackage{lipsum}
\usepackage{caption}
\usepackage{footnote}
\usepackage{tikz}
\usepackage[bottom]{footmisc}
\usepackage[toc,page]{appendix}

\newcommand{\ours}{Lynx\xspace}
\newcommand{\specialcell}[2][c]{%
  \begin{tabular}[#1]{@{}c@{}}#2\end{tabular}}

\title{What Matters in Training a GPT4-Style Language Model with Multimodal Inputs?}

\author{Yan Zeng\thanks{Equal Contribution. $^{\dagger}$Work done during an internship.}, ~~Hanbo Zhang$^{*}$, ~~Jiani Zheng$^{*\dagger}$, \\ \textbf{Jiangnan Xia},  ~~\textbf{Guoqiang Wei}, ~~\textbf{Yang Wei},  ~~\textbf{Yuchen Zhang}, ~~\textbf{Tao Kong} \\ \\
ByteDance Research \\
\href{https://lynx-llm.github.io}{\texttt{https://lynx-llm.github.io}}
}

\begin{document}

\maketitle

\begin{tikzpicture}[remember picture,overlay,shift={(current page.north west)}]
\node[anchor=north west,xshift=0.4cm,yshift=-3.0cm]{\scalebox{-1}[1]{\includegraphics[width=4cm]{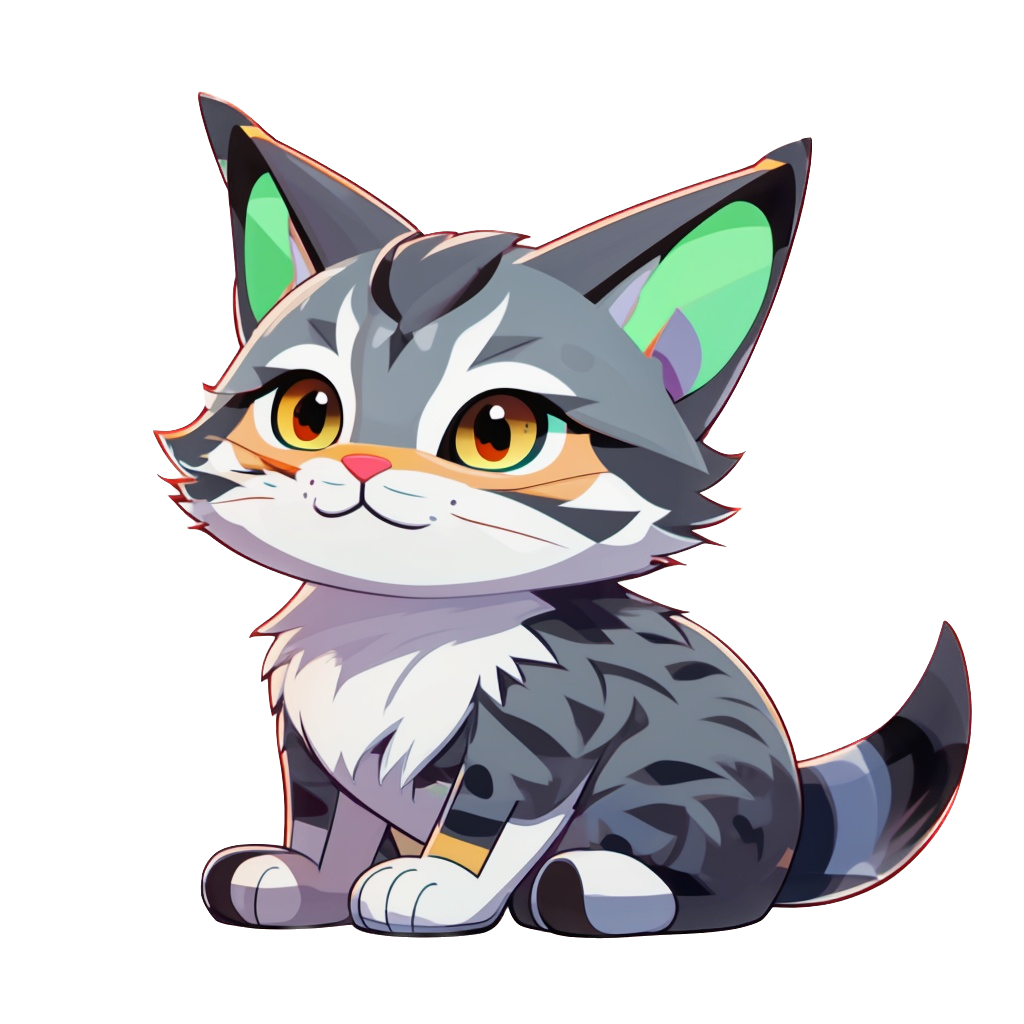}}};
\end{tikzpicture}

\begin{abstract}
Recent advancements in Large Language Models (LLMs) such as GPT4 have displayed exceptional multi-modal capabilities in following open-ended instructions given images. However, the performance of these models heavily relies on design choices such as network structures, training data, and training strategies, and these choices have not been extensively discussed in the literature, making it difficult to quantify progress in this field. To address this issue, this paper presents a systematic and comprehensive study, quantitatively and qualitatively, on training such models. We implement over 20 variants with controlled settings. Concretely, for network structures, we compare different LLM backbones and model designs. For training data, we investigate the impact of data and sampling strategies. For instructions, we explore the influence of diversified prompts on the instruction-following ability of the trained models. For benchmarks, we contribute the first, to our best knowledge, comprehensive evaluation set including both image and video tasks through crowd-sourcing. Based on our findings, we present \textbf{\ours}, which performs the most accurate multi-modal understanding while keeping the best multi-modal generation ability compared to existing open-sourced GPT4-style models. 
\end{abstract}

\section{Introduction}

Large Language Models (LLMs)~\cite{brown2020language, thoppilan2022lamda, hoffmann2022empirical, chung2022scaling, zhang2022opt, chowdhery2022palm, ouyang2022training, scao2022bloom, du2022glm, zeng2022glm, iyer2022opt, touvron2023llama, anil2023palm} have progressed rapidly in recent years and achieved impressive performance in language understanding and generalization.
With instruction fine-tuning \cite{ouyang2022training, chung2022scaling, wang2022self, vicuna2023, xu2023baize, peng2023instruction}, LLMs can be further improved to follow open-ended instructions from non-expert users and serve as dialog-based assistants in our daily lives. Leveraging powerful LLMs, recent studies have  examined methods for adapting LLMs to multimodal inputs (e.g., images \cite{li2023blip, alayrac2022flamingo, gong2023multimodal, liu2023visual, li2023otter, ye2023mplug, zhu2023minigpt, dai2023instructblip}, videos \cite{alayrac2022flamingo, li2023otter, chen2023videollm, maaz2023video, zhao2023learning, luo2023valley, zhang2023video}, and audio~\cite{zhang2023video, huang2023audiogpt}) and outputs (e.g., vision tasks~\cite{wang2023visionllm}, and robotic manipulation skills~\cite{jin2023alphablock, driess2023palm, jiang2022vima}).
Notably, GPT4 has astounded the world with its impressively stable zero-shot versatile yet practical capabilities, such as generating descriptions, stories, poetry, advertisements, and codes given images, which were rarely observed in previous vision language models~\cite{li2023blip, li2021align, bao2022vlmo, zeng2021multi, zeng2022x, li2022blip}.

However, it still remains a mystery that: \textit{How does GPT4 obtain its impressive smartness?} Though actively investigated recently, the existing models are usually different in network structure, training data, training recipes, prompts, and evaluation benchmarks, which makes it extremely hard to tell which factors are crucial in achieving a high-performance multi-modal language model. In addition, suitable quantitative benchmarks for evaluating and comparing such models are lacking, making it difficult to attribute and quantify the progress in open-sourced multi-modal LLMs.

% \begin{figure}
%     \centering
%     \includegraphics[width=\textwidth]{figs/demo.png}
%     \caption{Selected examples generated from Lynx.}
%     \label{fig:front-cases}
% \end{figure}

Therefore, in this paper, we conduct a systematic study on training GPT4-style models to address the aforementioned issues.
According to the existing literature, we identify three possible keys to achieving high performance for multi-modal LLMs: network structures, training data, and diversified instructions.
Regarding network structures, we explore different LLM adaptation strategies, including the widely utilized cross-attention-based structure~\cite{alayrac2022flamingo} and the recently popular decoder-only structure with a multi-modal adapter~\cite{liu2023visual, zhu2023minigpt, ye2023mplug}. 
Besides, we investigate different backbones including LLaMA-7B and Vicuna-7B to assess whether language instruction fine-tuning affects the final multi-modal performance. As for training data, we experiment with several large-scale datasets (e.g. COYO700M~\cite{kakaobrain2022coyo}, DataComp1B~\cite{gadre2023datacomp}, and BlipCapFilt~\cite{li2022blip}) consisting of image-text pairs to observe the effects of different data combinations. For instructions, we manually label at least three prompts for each task and generate more with GPT4 to figure out the influence of the diversity of language prompts. In total, \textit{there are 500 prompts for over 50 tasks}.
In summary, we implement $\sim$20 variants with controlled settings and conduct extensive experiments to draw reliable conclusions both quantitatively and qualitatively.

For benchmarking, we argue that the evaluation of multi-modal LLMs is essentially different from typical visual-language methods.
The primary challenge when evaluating a GPT4-style model is balancing text generation capability and multi-modal understanding accuracy. To address this, we present a new benchmark incorporating both video and image data to evaluate both the multi-modal understanding and text generation performances. Using our proposed benchmark, we evaluate a large bunch of open-source methods and provide a comprehensive review. Concretely, we adopt two protocols for quantitative evaluation. First, we collect an Open-ended Visual Question Answering (Open-VQA) test set, including questions on objects, OCR, counting, reasoning, action recognition, chronological ordering, and more. Different from standard VQA~\cite{antol2015vqa, zhang2016yin}, the ground-truth answer in Open-VQA is open-ended. To evaluate the performance on Open-VQA, we prompt GPT4 to make it a discriminator, yielding a 95\% agreement with human evaluation. This benchmark is used to evaluate the accuracy of all models. Additionally, we adopt the OwlEval test set proposed by mPLUG-owl~\cite{ye2023mplug} to assess the text generation ability given images. Though OwlEval is a tiny set containing only 82 questions based on 50 images, it covers a diverse range of tasks such as generating descriptions, stories, poems, advertisements, codes, and other sophisticated yet practical analyses of given images. In this part, we recruit human annotators to rank different models.

Based on extensive analysis of our controlled experiments, our findings can be summarized as follows:
\begin{itemize}
\item Prefix-tuning with trainable adaptors has shown better performances to adapt LLMs to multi-modal inputs compared to cross attention (e.g. Flamingo \cite{alayrac2022flamingo}).

\item Data quality is more important than quantity. We find that models trained on large-scale image text pairs like COYO700M and DataComp1B are not better to generate languages than models trained on a much smaller but high-quality dataset, since they can contaminate the output distribution. 

\item Diversified prompts are crucial to the improvement of the instruction-following ability and, consequently, final performance.

\item For the multi-modal adaptation of LLMs, it is crucial to carefully balance the multi-modal understanding and text generation abilities. Multi-modal adaptation based on instruction-finetuned models like Vicuna can improve the instruction-following abilities. 

\end{itemize}

Through our study, we present \textbf{\ours}, a simple prefix-tuning GPT4-style model, with a two-stage training recipe. 
For the first stage,  we use $\sim$120M image-text pairs to align visual and linguistic embeddings. For the second stage, we finetune our model with 20 multi-modal tasks with image or video inputs and NLP instruction data to learn to follow instructions. 
We transform all multi-modal datasets into the instruction-following format with manually written prompts and more GPT4-generated ones to keep the consistency of all training data. The resulting model performs the most accurate multi-modal understanding while exhibiting the best multi-modal generation ability compared to existing open-sourced models.

\section{\ours}

\ours is a GPT4-style large language model that can take images and videos as inputs.
Built on top of Vicuna, it is further trained with additional trainable adapters on high-quality image-text pairs and visual language tasks.
In this section, we will introduce our \ours in detail, including the problem formulation (\ref{sec:formulation}), architecture (\ref{sec:arch}), pretraining (\ref{sec:pretraining}), and instruction finetuning (\ref{sec:finetune}).

\begin{figure}
    \centering
    \includegraphics[width=\textwidth]{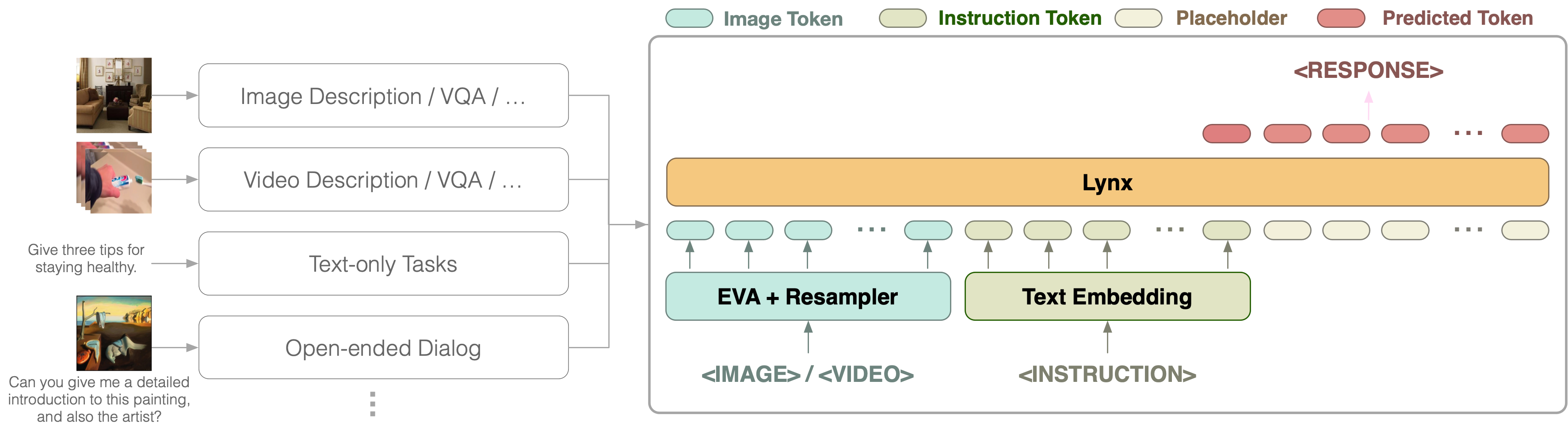}
    \caption{Our model is based on prefix-tuning architecture: the vision tokens are directly concatenated with the text tokens to generate outputs auto-regressively.}
    \label{fig:netarch}
\end{figure}

\subsection{Formulations}
\label{sec:formulation}
A GPT4-style large language model is defined as a decoder-only transformer \cite{radford2019language, brown2020language, openai2023gpt} that takes both visual and instructional tokens as inputs and generates responses in text auto-regressively.
Formally, the input includes vision tokens $\textbf{w}_v=\{w_i\}_{i=1}^{V}$ and instruction tokens $\textbf{w}_l=\{w_j\}_{j=V+1}^{V+L}$,
where $V$ and $L$ represent the number of vision tokens and instruction tokens. The vision tokens and instruction tokens in our model are directly concatenated to form the input of the decoder-only model. Conditioned on the multi-modal inputs, the model predicts the response in an auto-regressive manner, i.e., each word $w_i$ is predicted conditioned on all input tokens and previous predictions. Therefore, the sentence is predicted by the following equation:
\begin{equation}
    p(w_{V+L+1: V+L+T}|w_{1:V+L}) \sim \prod_{t=V+L+1}^{V+L+T}P(w_t | w_{< t})
\end{equation}
In large language models \cite{brown2020language, thoppilan2022lamda, hoffmann2022empirical, chung2022scaling, zhang2022opt, chowdhery2022palm, ouyang2022training, scao2022bloom, du2022glm, zeng2022glm, iyer2022opt, touvron2023llama, anil2023palm}, the network is usually trained on numerous text corpus to learn the causal relationships among tokens.
Similarly, our model is also trained on the collected visual-language tasks to learn the next-word distribution.
Notably, compared to the contrastive pretraining \cite{chen2020simple, radford2021learning}, pretraining with next-word prediction requires data with fluent texts that can represent the ``natural'' causal dependency between the predicted word and the past context very well \cite{brown2020language}.
We will introduce the details of data collection and selection in Section \ref{sec:pretraining} and \ref{sec:finetune} in detail.

\subsection{Details of Model Architecture} 
\label{sec:arch}

\paragraph{Overview}
\label{sec:prefix_tuning}
Our model takes simultaneously vision and language as inputs to generate text responses following the input instructions.
The overall structure of our model is shown in Fig.\ref{fig:netarch}.
Concretely, vision inputs are first processed by a vision encoder to get a sequence of vision tokens $\textbf{w}_v$.
After that, $\textbf{w}_v$ are fused with instruction tokens $\textbf{w}_l$ for multi-modal tasks.
In our model, we directly concatenate the projected vision tokens and instruction tokens as the input of LLMs, which can then be processed by the decoder-only LLMs naturally.
We call this structure ``prefix-finetuning'' (PT) in contrast to the cross-attention-based models like Flamingo \cite{alayrac2022flamingo}. Moreover, we find that by adding a small trainable adapter after some layers in the frozen LLMs, the performance could be further improved with low training costs.
To generate responses, the left-to-right causal decoder auto-regressively predicts the next token by taking all previous tokens as inputs until encountering the <EOS>.

\paragraph{Adapter}

\begin{wrapfigure}[13]{r}{0.6\textwidth}
    \centering
    \includegraphics[width=0.6\textwidth]{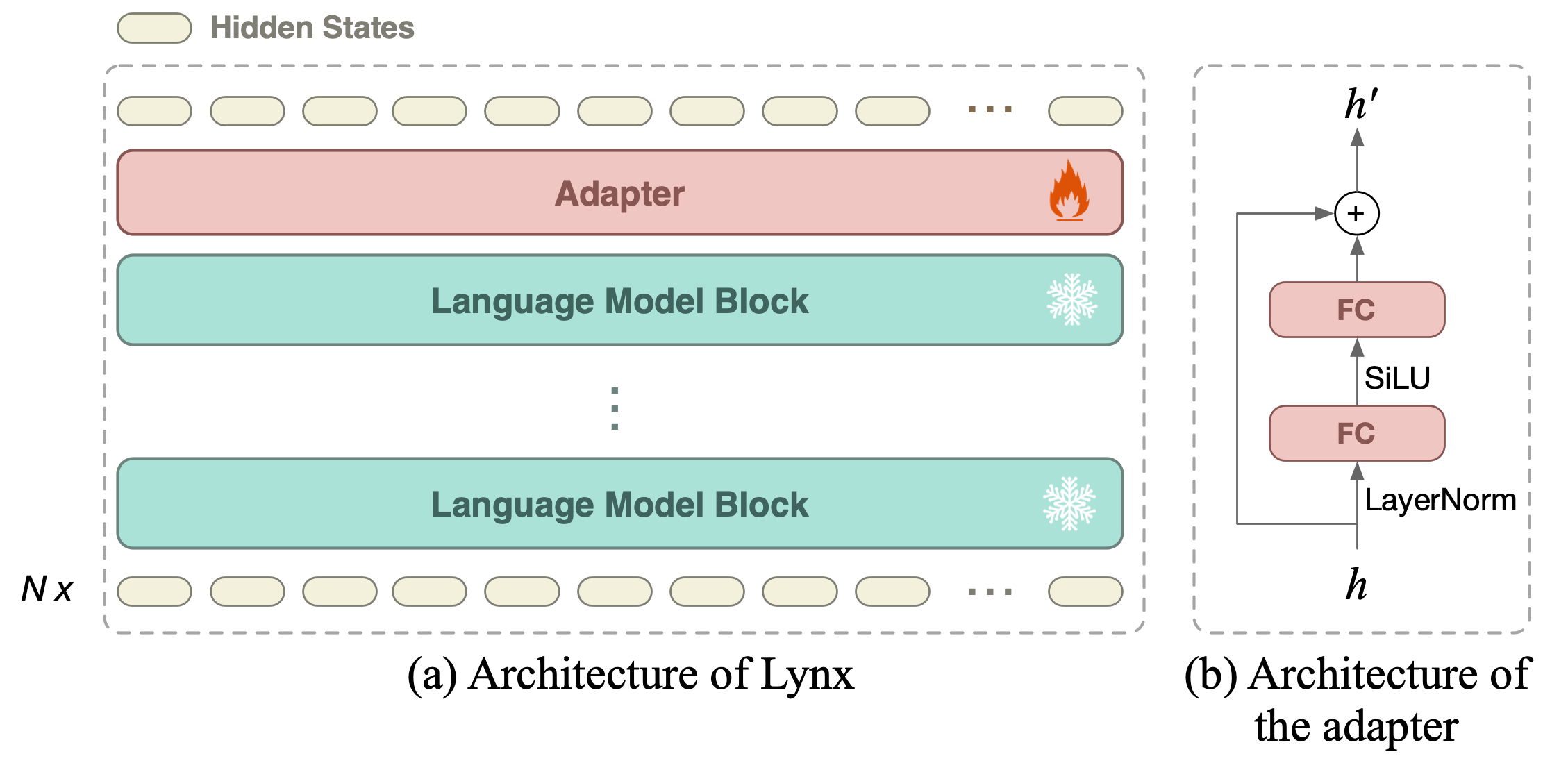}
    \caption{Architecture of \ours. (a) Overall; (b) Adapter.}
    \vspace{0pt}
    \label{fig:llamax}
\end{wrapfigure}

The trainable adapters are inserted into the LLMs after every $M$ blocks.
In our experiments, $M=1$. As shown in Figure \ref{fig:llamax}(b), the adapter linearly projects each token into a lower-dimensional space and then re-projects it back.
Concretely, in \ours, the hidden state for each token is 4096-d.
The adapter first imposes layer normalization \cite{ba2016layer} onto the hidden states.
Then a linear layer is used to downsample the dimension of each token state from 4096 to 2048, based on which SiLU \cite{elfwing2018sigmoid} is set as the non-linear activation function, which keeps consistent with LLaMA \cite{touvron2023llama}.
Finally, the other linear layer is used to re-map the 2048-d hidden state back to 4096-d.

\paragraph{Vision Encoder}
To extract vision features of images and video frames, we apply EVA-1B \cite{fang2023eva, sun2023eva} as our vision encoder $\phi_v(x)$.
It maps an image to a sequence of visual tokens.
The downsample rate is 14, meaning that an image with resolution $H\times W$ will be represented by a sequence of $\frac{H}{14}\times \frac{W}{14}$ tokens.
To improve the efficiency of training and inference, we adapt the resampler $\Phi$ mechanism \cite{jaegle2021perceiver, alayrac2022flamingo} that reduces the dimensions of vision inputs by injecting the long vision token sequence into a short and learnable query sequence $\textbf{w}_v^{q}$:
\begin{equation}
    \textbf{w}_v = \Phi(\phi_v(x), \textbf{w}_v^{q})
\end{equation}
where $x$ is the input image, $\phi_v(x)$ is the raw tokens directly given by the vision encoder, $\textbf{w}_v$ is the condensed token sequence consisting of 32 tokens regardless of the number of raw tokens from the vision encoder.

\subsection{Pretraining}
\label{sec:pretraining}

During pretraining, we utilize more than 120M image-text pairs to train the newly added layers so as to build connections of different modalities.
Our pretraining follows the typical next-word prediction training with the cross entropy loss.
To accelerate pretraining, we first pre-train our model on images of 224$\times$224 resolution.
Nevertheless, we found that only pretraining on a low resolution is not enough for some downstream tasks like table reading and OCR.
Therefore, after 100k steps of pretraining on low-res images, we continue to increase the input resolution to 420$\times$420 and train the model for another 10k steps.

Training data during this phase mainly consists of 
BlipCapFilt 115M \cite{li2022blip}, CC12M \cite{changpinyo2021conceptual}, CC3M \cite{sharma2018conceptual}, and SBU \cite{ordonez2011im2text}.
Besides, we also add high-quality labeled data during pretraining that have been also used in the instruction finetuning phase, like captioning, visual question answering, and classification.
Details of all pretraining datasets are listed in Table \ref{tab:ptdata}.
Our model is trained on a total of $\sim$14B tokens from all these datasets during the pretraining stage and $\sim$3B tokens during the instruction-finetuning stage.

\subsection{Instruction Fintuning}
\label{sec:finetune}
To finetune our model with diversified instructions, we collect an instruction finetuning multi-modal dataset based on the public ones.
Our dataset consists of 50+ text-only, image-text, and video-text tasks mainly belonging to 5 categories: Text-only Instruction-Following, Image/Video Visual Question Answering, Image/Video Captioning, Classification, and Image-conditioned Dialog for Complex Reasoning and Instruction Following.
We also provide the corresponding instructions for all of these tasks (see Appendix Table \ref{tab:ptdata} for details).
To do so, we manually labeled at least 3 different prompts for each of these tasks, and then invoke GPT4 to automatically generate more based on the following ``meta prompt'', i.e., the prompt used to generate prompts for different tasks:

\textit{
Here are some instructions that define a visual-language task. Continue to write 15 instructions with the same meaning: 1) PROMPT1; 2) PROMPT2; 3) PROMPT3;}

Besides, we also collect some available public (visual-)text instruction data (also listed in Table \ref{tab:ptdata}) to further improve the ability of our model to follow open-ended instructions, including the instruction data used in FlanT5 \cite{chung2022scaling}, Alpaca \cite{wang2022self}, Mini-GPT4 \cite{zhu2023minigpt}, LLAVA \cite{liu2023llava}, and Baize \cite{xu2023baize}. 

We follow the same causal prediction loss as in pretraining, i.e., the cross entropy loss to predict the next word based on all previous tokens.
Nevertheless, we observed that different weight combinations of the instruction data have a crucial influence on the final performance. 
Empirically, we finally impose the weight strategy presented in Table \ref{tab:ptdata}.

\section{Experiment}

In this section, we aim to answer the following questions according to empirical studies:

a) How can we evaluate the performance of a GPT4-style model? (Section \ref{sec:bench})

b) Compared to existing models, what are the advantages of our \ours? (Section \ref{sec:quantexp})

c) What matters to train a high-performance GPT4-style model? (Section \ref{sec:ablation})

d) What is the performance of \ours in open-world zero-shot scenarios? (Section Appendix \ref{sec:open-demo})

\subsection{Evaluation Protocols}
\label{sec:bench}

The evaluation of GPT4-style generative language models is challenging because the quality of natural languages is inherently subjective and highly depends on specific cases.
Existing models like PaLM-E \cite{driess2023palm}, PaLI \cite{chen2022pali}, BLIP2 \cite{li2023blip}, or InstructBLIP \cite{dai2023instructblip} turn to the evaluation on visual-language benchmarks like image caption \cite{chen2015microsoft} or visual question answering \cite{antol2015vqa}, i.e., fine-tuning multi-modal LLMs on a single downstream task on which the evaluation is conducted. 
Nevertheless, though it may achieve better performance, over-finetuning on such benchmarks will damage the generation ability of large language models, which conflicts with the primary motivation to use large language models.
Moreover, such benchmarks, especially the (semi-)automatically generated ones like TDIUC \cite{kafle2017analysis}, always contain a high ratio of easy or noisy examples, making them less suitable.
On the contrary, other methods like MiniGPT4 \cite{zhu2023minigpt} or LLaVA \cite{liu2023llava} only showcase their performance in some challenging yet practical scenarios without quantitative results due to the lack of quantitative benchmarks for such generative multi-modal language models.
Therefore, in this section, we propose to evaluate the GPT4-style models in the following two aspects:
\begin{itemize}
    \item A cleaned subset of visual-language benchmark, which should be challenging and compatible with generative models, with prompted GPT4 to get the quantitative results.
    \item An open-world challenging yet practical test set to evaluate the performance on realistic scenarios where GPT4-style models are needed, with humans to evaluate the user experience.
\end{itemize}

\begin{figure}
    \centering
    \includegraphics[width=\textwidth]{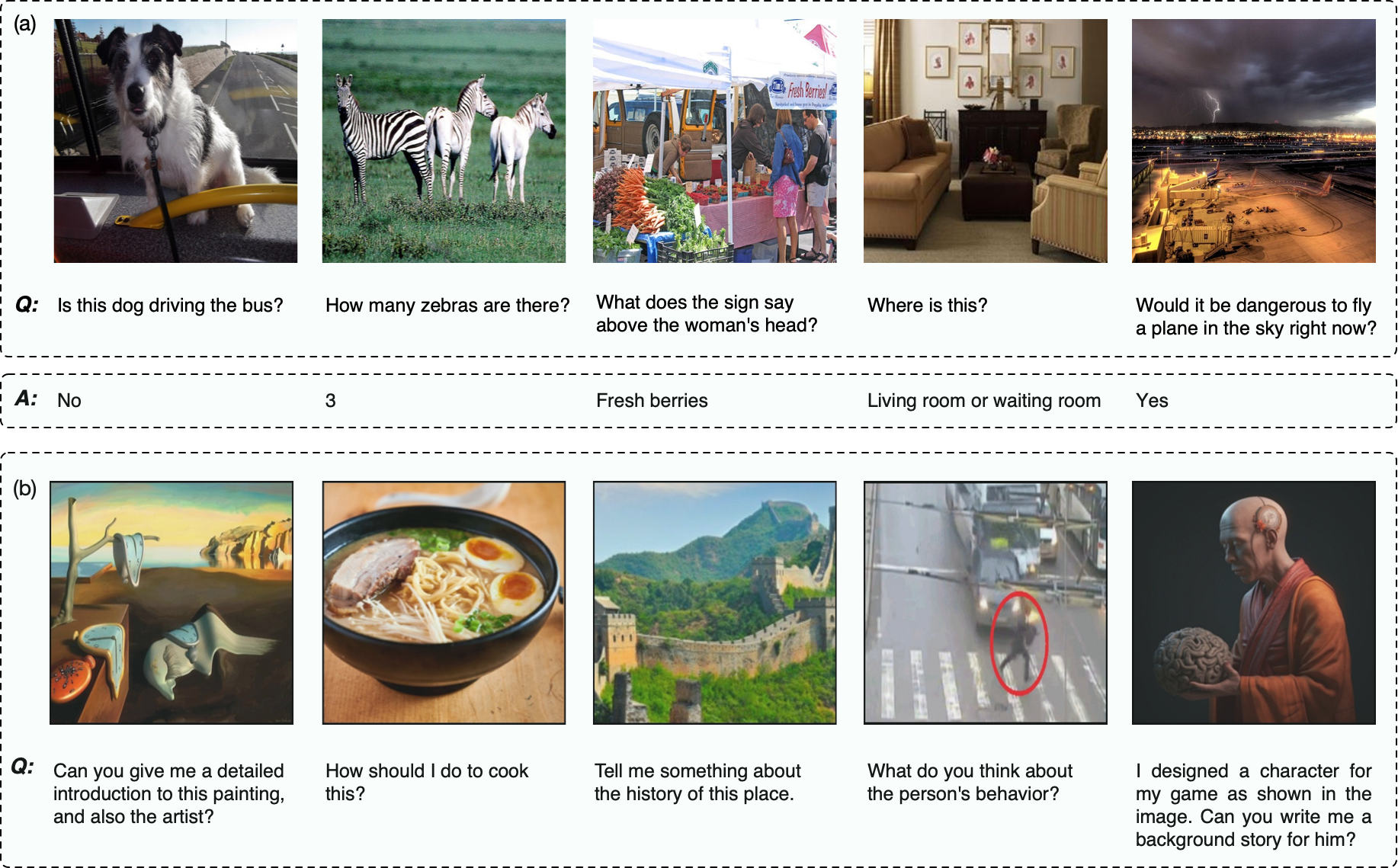}
    \caption{Examples of our test set. (a) Open-VQA benchmark to validate the accuracy of visual understanding; (b) OwlEval to evaluate the quality of language generation.}
    \label{fig:vqa}
\end{figure}

To do so, we manually collect an Open-VQA test set consisting of 450 samples with image or video input, which contains diverse questions on objects, OCR, counting, reasoning, action recognition, chronological ordering, etc., from VQA 2.0~\cite{antol2015vqa}, OCRVQA~\cite{mishra2019ocr}, Place365~\cite{zhou2017places}, MSVD~\cite{chen2011collecting}, MSRVTT~\cite{xu2016msr}, and Something-Something-V2 (SthV2)~\cite{goyal2017something}. 
Though Place365 is a classification task and SthV2 is a video captioning task, we write proper prompts to make them both VQA tasks.
Besides, we carefully examine the data and modify the questions and ground-truth answers if necessary to make them reliably correct and challenging enough to be a benchmark for GPT4-style models.
Randomly sampled examples are given in Fig.~\ref{fig:vqa}(a).
Different from the traditional VQA benchmark, Open-VQA supports open-ended answers.
To achieve so, we prompt GPT4 to make it the referee, which achieves a consistency of more than 95\% compared with humans\footnote{We evaluate the consistency on 100 samples from a randomly selected subset with our model.}.
The prompt for GPT4 used in this phase is as follows:

\textit{Given the question ``QUESTION'', does the answer ``PREDICTION'' imply the answer ``GROUND\_TRUTH''? Answer with Yes or No.}

Moreover, general-purpose language generation with image inputs is also important to multi-modal LLMs. 
Therefore, we also adopt the OwlEval test set proposed by mPLUG-owl \cite{ye2023mplug}, which contains 82 questions based on 50 images, where 21 from MiniGPT-4 \cite{zhu2023minigpt}, 13 from MM-REACT~\cite{yang2023mmreact}, 9 from BLIP2 \cite{li2023blip}, 3 from GPT4 \cite{openai2023gpt}, and 4 collected by mPLUG-owl itself.
The test set includes diversified and practical cases such as dense image captioning, dialogue writing, story writing, poem writing, teaching, programming, etc.

We give some examples in Fig.\ref{fig:vqa}(b).
However, OwlEval is proposed together with mPLUG-owl.
Hence, directly using it as the benchmark is possibly unfair to other models.
To make the comparison fair, we pad each image in the OwlEval with 8 pixels as shown in Fig.\ref{fig:vqa}(b) before feeding them into the models.
We recruit human annotators to evaluate the performance.
Scores range from 1 to 5.
If two models are considered to be equally good or bad, they will have the same score.
For each data, the annotator will assign a score for each model.
We only allow at most 2 models that are equally good or bad, and for each annotator, the total number of ties should be no more than 10 for the whole set.
During the evaluation, the correctness has the highest priority, then should be the richness of the generated content.

Finally, we also compare our method with others on the newly proposed MME benchmark \cite{fu2023mme}, which includes 14 different subtasks that evaluate the perception and cognition ability of multi-modal large language models.

\begin{table*}[t]
\begin{center}
% \small 
\resizebox{\linewidth}{!}{
\begin{tabular}{l|cccccccc|c}
\toprule 
 & OCR & Counting & Reasoning & Place & Color & Spatial & Action & Others & Overall \\
\midrule

Open-Flamingo-0 & 20/53 & 5/37 & 15/31 & \textbf{18/22} & 5/30 & 7/15 & 11/20 & 53/94 & 44.37 \\

Open-Flamingo-4 & 14/53 & 6/37 & 15/31 & 17/22 & 9/30 & 7/15 & 11/20 & 51/94 & 43.05 \\

Multimodal GPT & 19/53 & 8/37 & 21/31 & 12/22 & 8/30 & 6/15 & 12/20 & 56/94 & 47.02\\

MiniGPT-4 & 32/53 & 13/37 & 13/31 & 17/22 & 16/30 & \textbf{9/15} & 16/20 & 63/94 & 59.27\\

LLaVA & 21/53 & 8/37 & 13/31 & 11/22 & 12/30 & 4/15 & 16/20 & 49/94 & 44.37 \\ 

mPLUG-owl & 34/53 & 8/37 & 16/31 & 16/22 & 14/30 & \textbf{9/15} & 13/20 & 62/94 & 56.95 \\

% use pretrain_flant5xl follow the blip2/eval/vqav2 default setting
BLIP2 & 29/53 & 15/37 & 21/31 & 12/22 & 17/30 & 8/15 & 16/20 & 67/94 &  61.26 \\ 

% blip2_vicuna_instruct vicuna7b
InstructBLIP & \textbf{41/53} & 20/37 & \textbf{26/31} & 14/22 & \textbf{23/30} & 6/15 & \textbf{18/20} & 77/94 & 74.50 \\ 

\midrule

Ours & 36/53 & \textbf{25/37} & \textbf{26/31} & 17/22 & 21/30 & \textbf{9/15} & 17/20 & \textbf{79/94} & \textbf{76.16}  \\ 

\bottomrule
\end{tabular}}
\end{center}
\caption[caption]{Comparison of existing open-sourced multi-modal LLMs and quantitative evaluation results (accuracy) on our Open-VQA image test set. For all models, we apply the same hyper-parameters defined in Appendix~\ref{app:hyper}. }
\label{tab:vqares}
\end{table*}

\begin{table}[t]
\begin{minipage}[b]{0.45\linewidth}
\centering
\resizebox{\linewidth}{!}{
\begin{tabular}{l|cc|c}
\toprule 
 & Action (Y/N) & Others & Overall \\
\midrule

InstructBLIP & 62/108 & 21/40 & 56.08 \\ 

mPLUG-owl & 65/108 & 19/40 & 56.76\\
MiniGPT-4 & 56/108 & 18/40 & 50.00\\
\midrule

Ours & 69/108 & 29/40 & 66.22 \\ 
\bottomrule
\end{tabular}}
\vspace{10pt}
\caption{Comparison of existing open-sourced multi-modal LLMs on the Open-VQA video benchmark.}
\label{tab:video-vqa}
\end{minipage}\hfill
\begin{minipage}[b]{0.5\linewidth}
\centering
\includegraphics[width=0.7\linewidth]{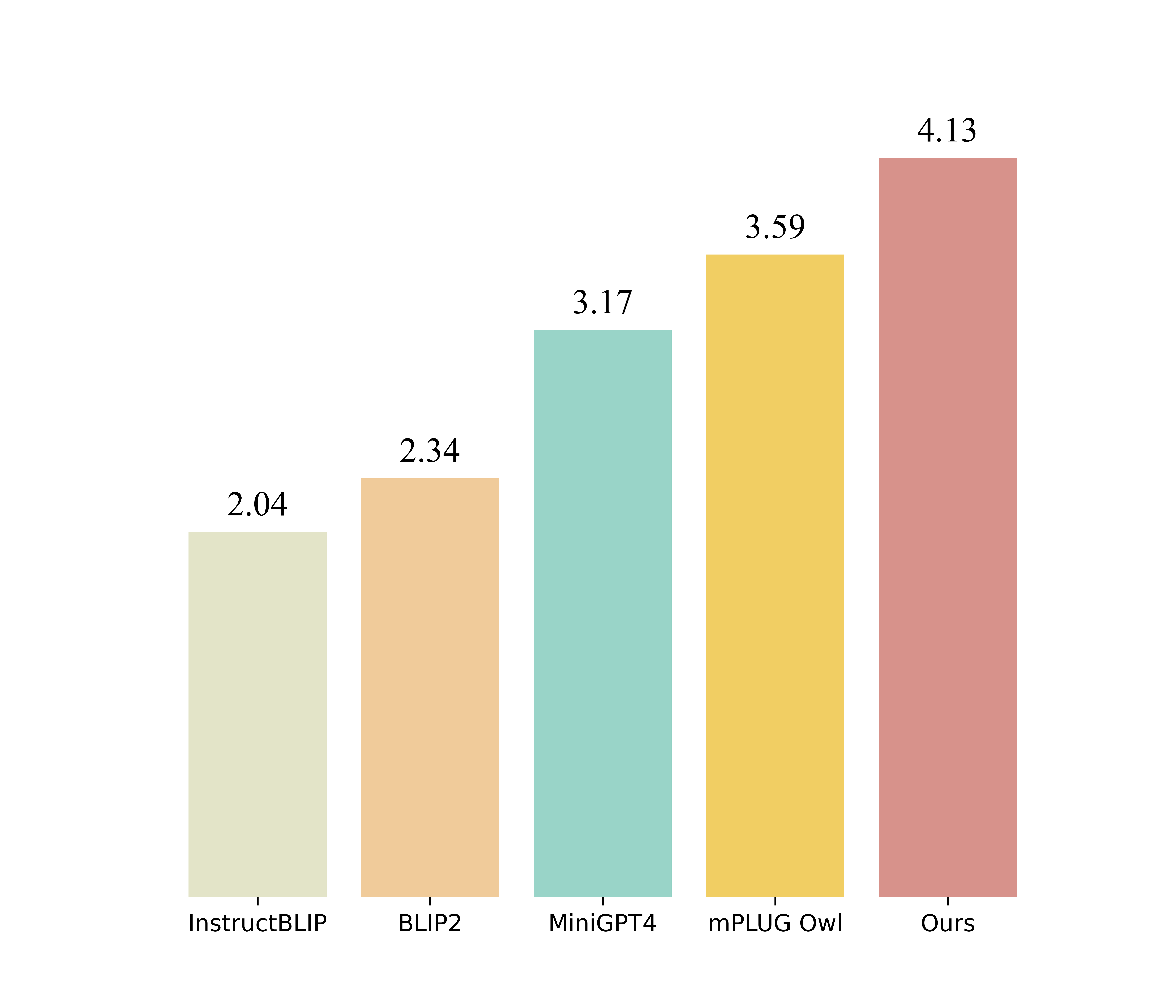}
\captionof{figure}{Comparison of human-evaluation performance on OwlEval. Scores are averaged over the number of questions.}
\label{fig:owleval}
\end{minipage}
\end{table}

\subsection{Quantitative Experiments}
\label{sec:quantexp}

\begin{figure}
    \centering
    \includegraphics[width=\textwidth]{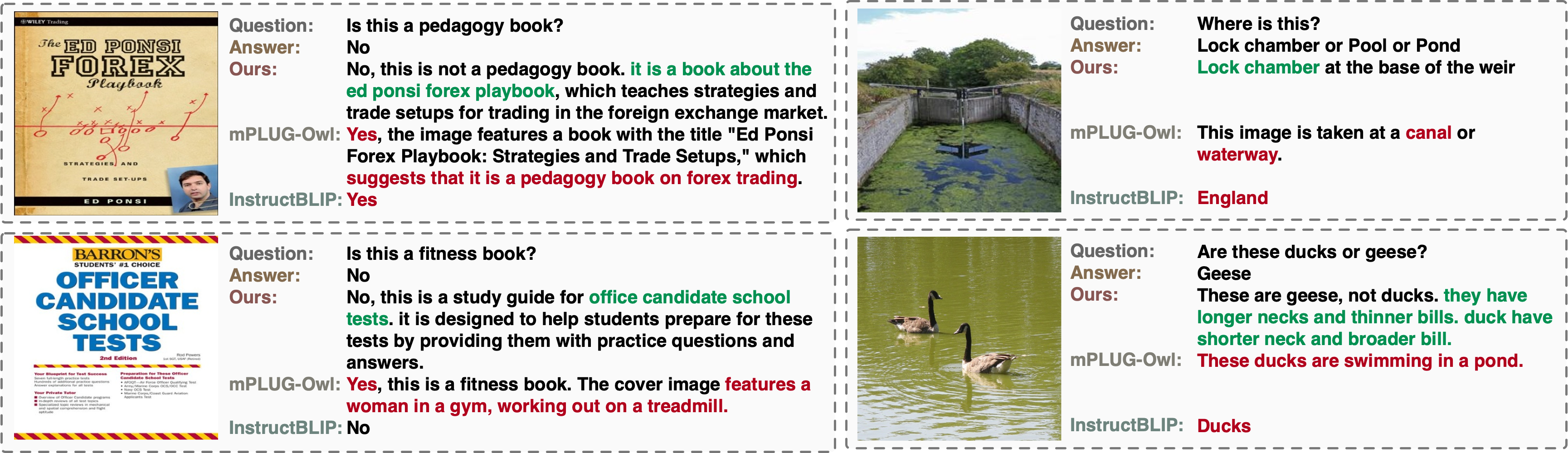}
    \caption{Qualitative results on our Open-VQA benchmark of different models. We choose InstructBLIP and mPLUG-Owl because they perform best on the Open-VQA benchmark and OwlEval benchmark in all baseline algorithms.}
    \label{fig:vqa-cases}
\end{figure}

\begin{figure}
    \centering
    \includegraphics[width=\textwidth]{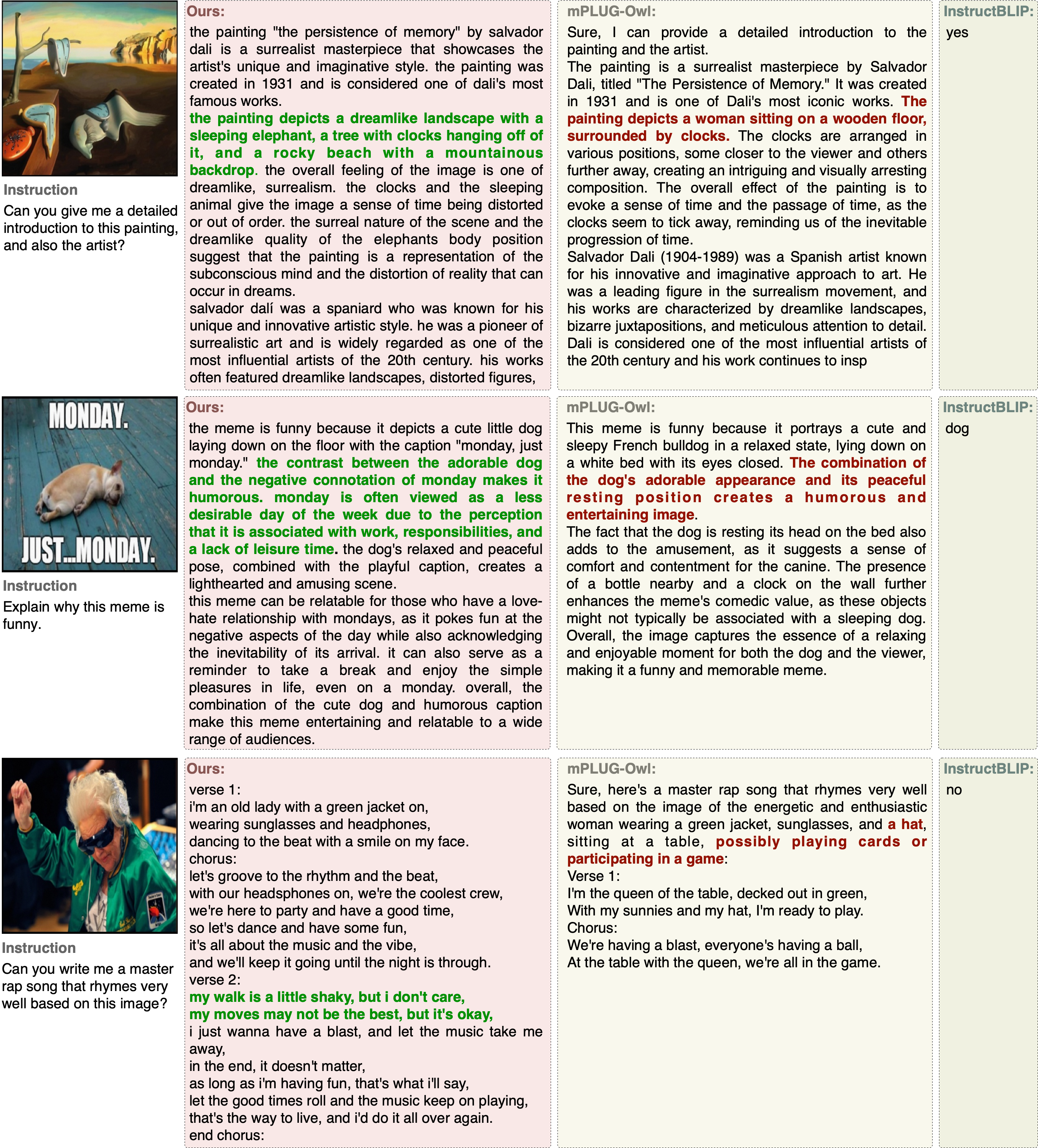}
    \caption{Qualitative results on OwlEval benchmark of different models. We choose InstructBLIP and mPLUG-Owl because they perform best on the Open-VQA benchmark and OwlEval benchmark in all baseline algorithms.}
    \label{fig:owl-cases}
\end{figure}

\paragraph{Open-VQA benchmark} We first evaluate our model as well as several existing open-sourced multi-modal LLMs on the Open-VQA benchmark.
Results are shown in Table~\ref{tab:vqares}.
We can conclude that our model has achieved the best performance both in the image and video understanding tasks.
Notably, InstructBLIP \cite{dai2023instructblip} also achieves high performance in most cases, even better than our model in OCR, color recognition, and action recognition tasks.
However, we observe that it always outputs one word for the question as shown in Fig.\ref{fig:vqa-cases} and \ref{fig:owl-cases}, which is less preferred by most of the users (see Fig.\ref{fig:owleval}).
We also showcase some of the examples in Fig.~\ref{fig:vqa-cases}.
More cases including video VQA examples can be found in Fig.~\ref{fig:more-image-vqa-cases} and \ref{fig:more-video-vqa-cases} in the appendix.
We can see that our model can give the correct answer in most cases as well as a concise reason that supports the answer, which makes it more user-friendly.

\paragraph{OwlEval benchmark} 
We evaluate the performances of general-purpose natural language generation on OwlEval test set. 
From the human evaluation results in Fig.\ref{fig:owleval}, we can see that our model has the best language generation performance while keeping high performance on the Open-VQA benchmark.
BLIP2 \cite{li2023blip} and InstructBLIP \cite{dai2023instructblip}, though achieved high performance on the Open-VQA benchmark, are not preferred by human users due to their extremely short outputs, i.e., in most cases, they only output one word or phrase as the answer without any explanation.
In contrast, MiniGPT4 \cite{zhu2023minigpt} and mPLUG-Owl \cite{ye2023mplug} are trained less to fit the Open-VQA benchmark and keep more language generation ability.
Hence, they are preferred over the BLIP models, though they may make more factual errors.
We also show some results on the OwlEval in Fig.~\ref{fig:owl-cases}.

In general, we observe that if a model has lower accuracy on the Open-VQA benchmark, it tends to make factual errors inconsistent with the given image during text generation.
Nevertheless, models with higher performance on the Open-VQA benchmark usually tend to lose language generation ability, e.g., generate short sentences.
We attribute this conclusion to the under-training or over-training on visual-language tasks.
To be specific, existing training data from visual-language tasks always includes short outputs.
By training on these data, the model can learn to align the visual and linguistic concepts, yet lose the language generation ability inherited from the large language model.
From the high performance of our model, we can see that one possible way to train a high-performance model with better language generation ability is to carefully select and clean the data, as well as design the proper sampling ratios.
Nevertheless, the key to balance language generation and correctness is a high-quality visual-language dataset that contains clean and rich expressions, which should be explored in our future work.

\paragraph{MME benchmark}

\begin{wrapfigure}[22]{r}{0.6\textwidth}
    \centering
    \includegraphics[width=0.6\textwidth]{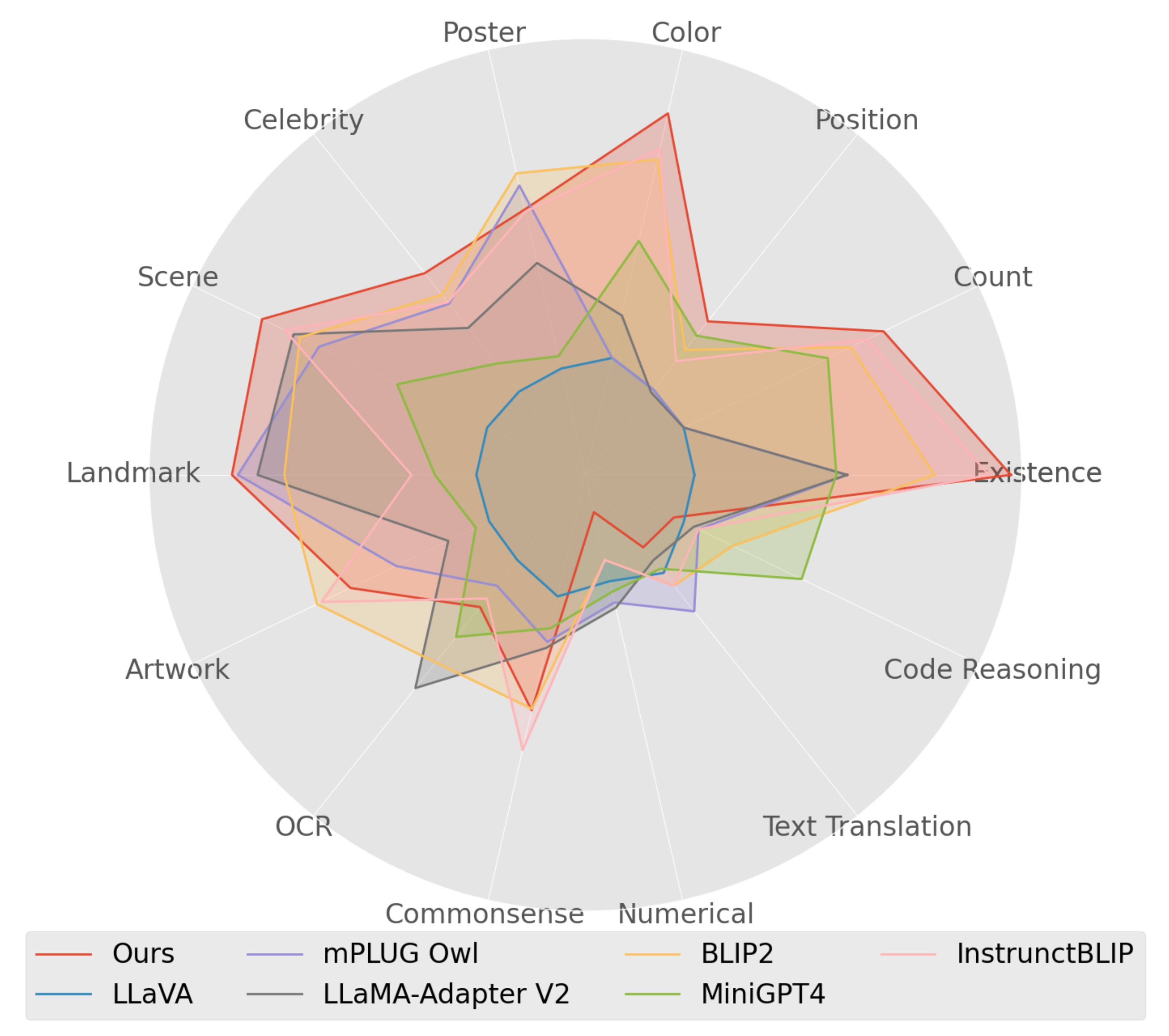}
    \caption{Comparison on MME benchmark.}
    \vspace{0pt}
    \label{fig:mme}
\end{wrapfigure}

We also compare \ours with available existing open-source models on the MME benchmark \cite{fu2023mme}.
Results are shown in Figure \ref{fig:mme} and Appendix \ref{ap:mme}.
We can see that our model is a state-of-the-art model in 7 out of 14 subtasks, especially for the perception tasks including Color, Celebrity, Scene, Landmark, Position, Count, and Existence.
Yet, from the figure, we can also see that our model seems not to perform well on cognition tasks including Code Reasoning, Text Translation, and Numerical.
Notably, cognition benchmarks including Code Reasoning, Text Translation, and Numerical in MME only contain 20 examples, which may cause high variance in the evaluation of different checkpoints.

\subsection{Ablation Study}
\label{sec:ablation}

We conduct an in-depth ablation study to investigate the impact of different components or training recipes on multi-modal understanding and language generation performances.
In this section, we follow the same evaluation method proposed in Section \ref{sec:bench}.

\begin{table*}[t] 
\begin{center}
\resizebox{\linewidth}{!}{
\begin{tabular}{l|cccccccc|c}
\toprule 
 & OCR & Counting & Reasoning & Place & Color & Spatial & Action & Others & Overall \\

\midrule

w/ LLaMA & 33/53 & 18/37 & 19/31 & 17/22 & 22/30 & 10/15 & 17/20 & 78/94 & 70.86 \\ 

w/o diverse prompts & 33/53 & 22/37 & 23/31 & 20/22 & 21/30 & 12/15 & 17/20 & 80/94 & 75.50 \\

w/ large-scale noisy data & 33/53 & 20/37 & 28/31 & 17/22 & 17/30 & 10/15 & 16/20 & 79/94 & 72.85 \\

w/ cross-attn & 13/53 & 6/37 & 11/31 & 3/22 & 8/30 & 9/15 & 5/20 & 41/94 & 31.79 \\ 
w/ cross-attn \& trainable LLM  & 26/53 & 15/37 & 28/31 & 14/22 & 17/30 & 8/15 & 14/20 & 63/94 & 61.26 \\ 

w/o high-resolution & 30/53 & 20/37 & 26/31 & 15/22 & 25/30 & 8/15 & 19/20 & 79/94 & 73.51 \\ 
\midrule
Ours & 36/53 & 25/37 & 26/31 & 17/22 & 21/30 & 9/15 & 17/20 & 79/94 & 76.16 \\ 
\bottomrule
\end{tabular}}
\end{center}
\caption[caption]{Ablation study on our Open-VQA images.}
\label{tab:image-ab}
% \caption{}
\end{table*}

\begin{table}[t]
\begin{minipage}[b]{0.53\linewidth}
\centering
\resizebox{\linewidth}{!}{
\begin{tabular}{l|cc|c}
\toprule 
 & Action (Y/N) & Others & Overall  \\
\midrule
w/ LLaMA & 65/109 & 25/40 & 60.81 \\
w/o diverse prompts & 62/109 & 26/40 & 59.46 \\ 
w/ large-scale noisy data & 63/109 & 26/40 & 60.14\\
w/ cross-attn & 63/109 & 13/40 & 51.35\\ 
w/ cross-attn, tune LLM & 59/109 & 19/40 & 52.70 \\ 
w/o high-resolution & 66/109 & 26/40 & 62.16\\ 
\midrule
Ours &  69/109 & 29/40 & 66.22\\
\bottomrule
\end{tabular}}
\vspace{10pt}
\caption{Ablation study on our Open-VQA videos.}
\label{tab:video-ab}
\end{minipage}\hfill
\begin{minipage}[b]{0.45\linewidth}
    \centering
    \includegraphics[width=\linewidth]{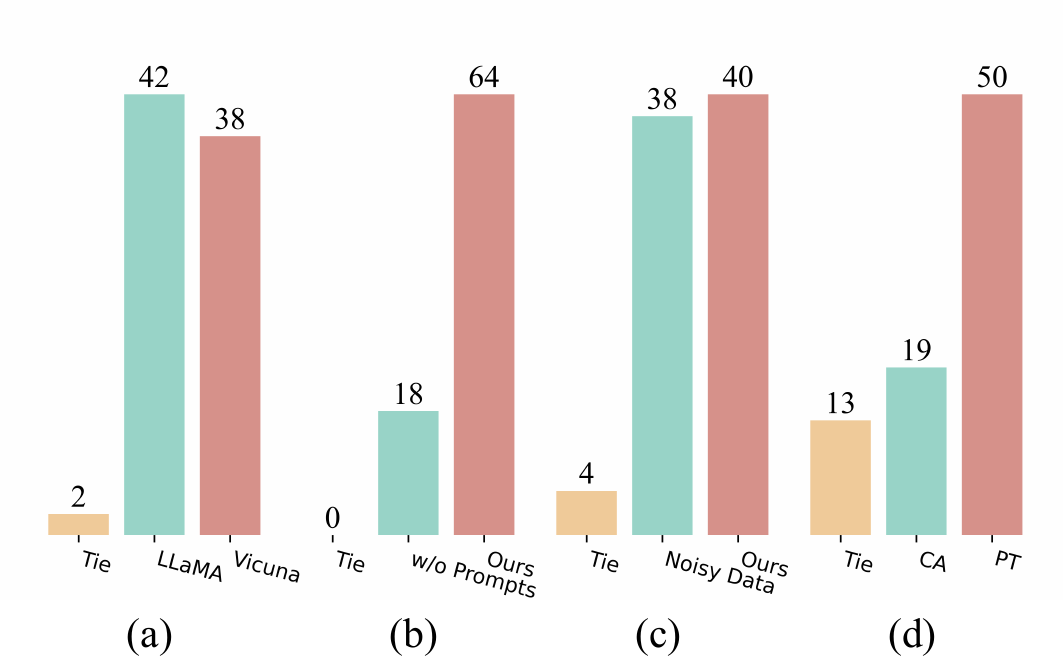}
    \captionof{figure}{Human evaluation of different ablation models. (a) w/ LLaMA vs w/ Vicuda; (b) w/o diversified prompts vs w/ diversified prompts; (c) w/ large-scale noisy data vs w/o large-scale noisy data; (d) prefix-finetuning vs cross-attention.}
    \label{fig:human-eval-ab}
\end{minipage}
\end{table}

\begin{figure}
    \centering
    \includegraphics[width=\textwidth]{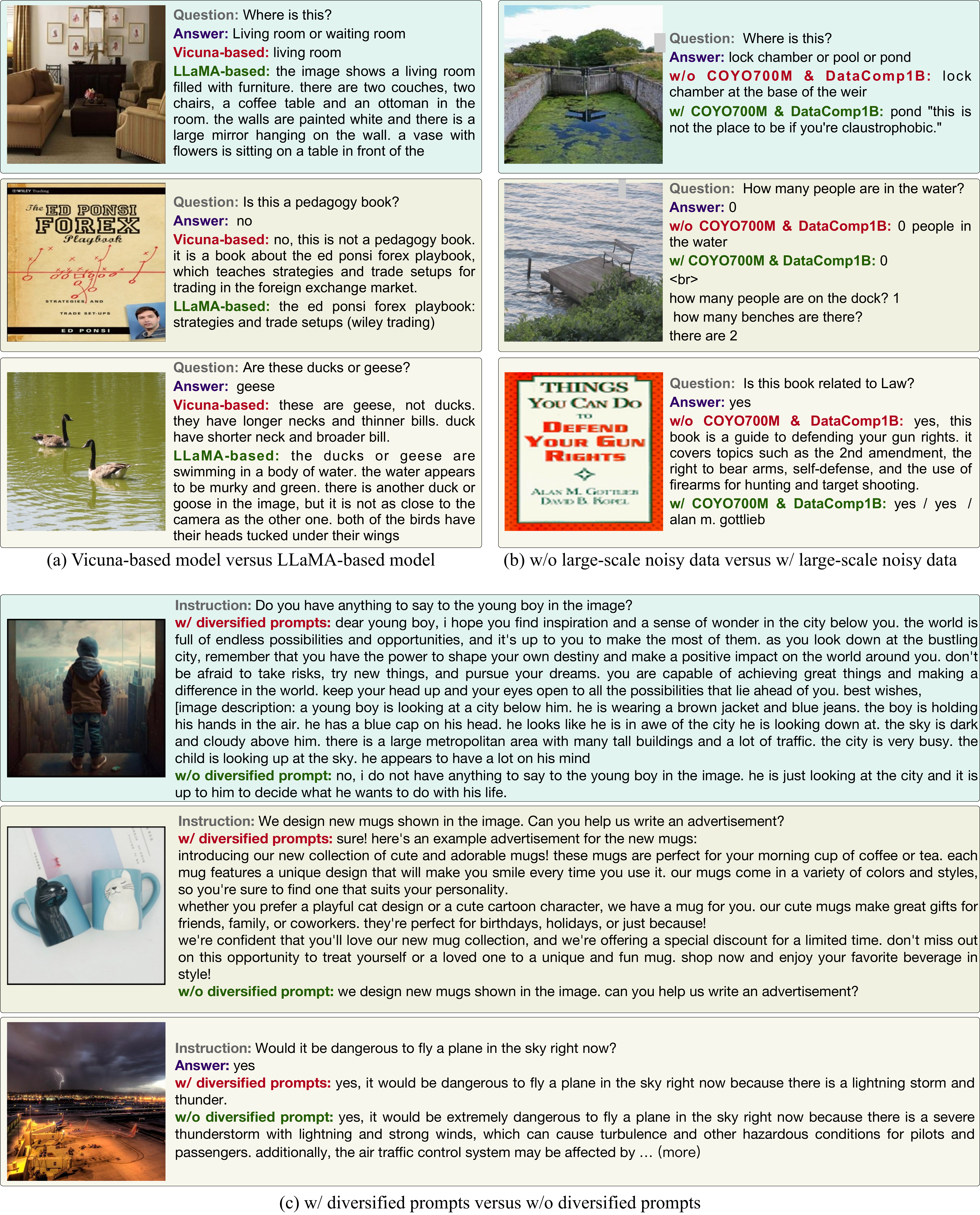}
    \caption{Ablation study cases on (a) Vicuna-based model versus LLaMA-based model; (b) w/o large-scale noisy data versus w/ large-scale noisy data; (c) w/ diversified prompts versus w/o diversified prompts.}
    \label{fig:ab-cases}
\end{figure}

\paragraph{LLaMA vs. Vicuna} 
As shown in Table \ref{tab:image-ab}, our experiments show that in the aspect of correctness, instruction-finetuned backbone (e.g.Vicuna) performs slightly better on our Open-VQA benchmark (like LLaVA) as shown in Table \ref{tab:image-ab} and \ref{tab:video-ab}, but slightly worse on the OwlEval benchmark (Figure \ref{fig:owleval}).
However, Vicuna-based model does indeed follow the instruction better. For example, the average answer length given the instruction ``give a short answer'' is 15.81, compared to 20.15 from the LLaMA-based model. One can also refer to Figure \ref{fig:ab-cases}(a) for examples of the comparison in terms of their instruction-following ability.

\paragraph{Impact of Diversified Prompts} 
It has been proved to be important to train LLMs on instruction data so as to make them follow instructions correctly \cite{chung2022scaling, ouyang2022training}.
Therefore, we ablate our model with diversified prompts written by both users and GPT4.
The results in Table~\ref{tab:image-ab} and \ref{tab:video-ab} show that our prompts help to balance different abilities.
Moreover, we also find that by using diversified prompts, our model can follow the open-ended instructions better than the ones trained without these prompts (Table \ref{tab:prompts}).
This observation accords with the text-only models.
The human evaluation results in Figure \ref{fig:human-eval-ab}(b) also accord with our observations.
Diversified tasks and prompts will help to improve the generalization of the model to new tasks and instructions.

\paragraph{Impact of Training Data}
We investigate the impact of data quantity and quality by training our model with or without the large-scale yet noisy image-text pairs (COYO700M \cite{kakaobrain2022coyo} and DataComp1B \cite{gadre2023datacomp}).
During our experiments, we find training data in both pretraining and finetuning largely influence the model performance.
Different from traditional visual-language pretraining \cite{radford2021learning}, we find that multi-modal LLMs do not benefit from large-scale but noisy image-text pairs because many of the texts in such datasets are not fluent or natural language expressions. 
For the generative pretraining in our model, they largely damage the language generation ability as shown in Figure \ref{fig:ab-cases}(b).
As a result, pretraining on such large-scale datasets achieves no better results than only training on a much smaller but cleaner dataset as evaluated by the human users as shown in Figure \ref{fig:human-eval-ab}(c).

\paragraph{Prefix-Tuning vs. Cross-Attn} We follow Flamingo \cite{alayrac2022flamingo}, concretely Open-Flamingo \cite{anas_awadalla_2023_7733589}, to implement the cross-attention method. 
Following its original settings, we only use multi-modal instruction data for pre-training.
For the finetuning stage, we experiment with two variants, with or without trainable LLM, i.e., with or without the use of text instruction data.
As shown in Table \ref{tab:image-ab} and \ref{tab:video-ab}, both of them perform worse than our prefix-tuning with adapters.
Though the models can generate fluent and relevant responses, their outputs usually do not give correct answers to the questions.
We also verified our conclusion with human annotators, as shown in Figure \ref{fig:human-eval-ab}(d).
Results show that human users give lower preference to the cross-attention models. Overall, cross-attention models could require more hyper-parameter searching to achieve better performances, and we leave it to further work.

\paragraph{Impact of Larger Image Resolution} 
We increase image resolution in the first stage with only 10K step training.
After that, we freeze the vision encoder and thus the expense of increasing image resolution is affordable.
For rigor, we also conducted an experiment to verify the impact of image resolutions on the model performance.
The experiment results in Table \ref{tab:image-ab} and \ref{tab:video-ab} show that the training on 420x420 resolution achieves better performance than the models only trained on 224x224.

\section{Related Work}

\paragraph{Large-language models.}
Large language models (LLMs) have been widely investigated in recent years due to their good generality on zero-shot tasks, including GPT3 \cite{brown2020language}, PaLM \cite{chowdhery2022palm, anil2023palm}, BLOOM \cite{scao2022bloom}, Chinchilla \cite{hoffmann2022training}, T5 \cite{raffel2020exploring}, LLaMA \cite{touvron2023llama}, OPT \cite{zhang2022opt}, GLM \cite{du2022glm}, etc.
After being pre-trained on massive text corpora, such models can perform surprisingly well on downstream tasks without further finetuning.
In particular, the simple yet efficient structure of decoder-only models like GPT-3 can easily scale up to hundreds of billions of parameters and show an elegant scaling law with the increase of model size and data amounts \cite{kaplan2020scaling}.
Moreover, recent advances in instruction finetuning \cite{iyer2022opt, chung2022scaling, ouyang2022training} have also shown that large-scale language models can be finetuned with limited amounts of instruction data to follow open-ended instructions in natural language.
This not only improves their performance on downstream tasks substantially but also makes it a user-friendly assistant in our daily life \cite{openai2023gpt}.

\paragraph{Centralized Multi-modal Interactive System.} Inspired by the achievements in LLMs, it is straightforward to ask a question: \textit{Is it possible to design a model that accepts multi-modal inputs while being able to chat with humans in natural language?}
Therefore, recent works investigate actively to design of such multi-modal interactive models.
One of the most intuitive ideas, such as Visual ChatGPT~\cite{wu2023visual}, MM-REACT~\cite{yang2023mm}, HuggingGPT~\cite{shen2023hugginggpt}, InternGPT~\cite{liu2023internchat}, SayCan~\cite{ahn2022can}, InnerMonologue~\cite{huang2022inner}, integrates various existing individual models or tools such as OCR, object detection, image captioning, visual question answering, text-to-image generation, or robot manipulation policies by a centralized controller.
In such a system, the LLM works as a ``manager'' that directly accepts instructions from users and selects the most appropriate tools to respond to requests while the integrated individual models are ``workers'' responsible for a specific kind of task.
Typically, such models are powerful to address problems that are already well-defined.
Yet, they, to some extent, lack zero-shot ability when encountering open-ended instructions which cannot be handled by any of their workers.

\paragraph{End-to-end Multi-modal Large Language Models.} By contrast, inspired by the recent advances of LLMs, it has also been shown feasible and promising to directly train the neural networks that directly accept multi-modal inputs and output responses end-to-end.
To achieve so, one intuitive idea is to adapt the LLMs to multi-modal inputs by adding some additional trainable parameters and finetuning them on multi-modal data.
For example, Flamingos \cite{alayrac2022flamingo} is one of the early works to explore this idea.
Firstly, it takes a vision encoder (like NFNet \cite{brock2021high} in their original version, or recent CLIP ViT \cite{radford2021learning}) to extract visual embeddings.
Then, it applies multi-layer cross-attention to fuse the multi-modal inputs for the final prediction.
Recent works directly concatenate vision embeddings to the inputs of LLMs and finetune LLMs end-to-end.
To do so, they usually add an additional projection layer to map the vision embeddings to the same dimension as the language embeddings, and then directly feed them into LLMs for further training.
Different methods may take different training strategies.
BLIP2 \cite{li2022blip} designs a Q-Former, which is the only trainable part, to align the dimensions of vision and language tokens.
PaLM-E \cite{driess2023palm}, which is built upon PaLM \cite{chowdhery2022palm}, is trained totally end-to-end with no fixed layers using a mix of multi-modal datasets including WebLI 10B dataset \cite{chen2022pali}.
Mini-GPT4 \cite{zhu2023minigpt} freezes all weights of the vision encoder and the LLM while only finetuning the weights of the projection layer.
LLAVA \cite{liu2023llava} fixes the vision encoder while keeping the LLMs trainable during the instruction finetuning stage.
mPLUG-owl \cite{ye2023mplug} tunes the vision encoder and keeps LLMs fixed to align the vision and language embeddings in the first stage while further tuning the LLMs and keeping the vision encoder fixed in the second instruction-finetuning stage.
KOSMOS-1 \cite{huang2023language} does not rely on any pretrained LLMs and is trained from scratch on large amounts of mixed data including image-text pairs (COYO700M \cite{kakaobrain2022coyo}, LAION2B \cite{schuhmann2022laion}, etc.), text corpora (Common Crawl, the Pile \cite{pile}, etc.), and interleaved image-text data.
These models are all powerful and show promising results to develop multi-modal large language models.

\section{Discussions and Limitations}

\subsection{Findings and Takeaways}

\paragraph{Prefix-tuning has shown better performances than cross-attention methods on multi-modal adaptation for large language models.} 
As shown in our experiments, prefix-tuning with adaptors show good performance on open-ended instruction-following tasks after training in billions of multi-modal tokens. By contrast, cross-attention models are not that efficient to achieve good performance, though more hyper-parameter searching could improve its performances and we leave it in future work.

\paragraph{Multi-modal LLMs are not as instruction-following as LLMs.}
In our experiments, we find that current multi-modal LLMs are not as good at the instruction following as language models.
For example, InstructBLIP \cite{dai2023instructblip} tends to generate short responses regardless of the input instructions, while other models tend to generate long sentences without considering the instruction like ``Give a short answer'' or ``Answer in one word''.
We assume that this is from the lacking of high-quality and diversified multi-modal instruction data.

\paragraph{The quality of training data is critical to model performance.}
As concluded in Section \ref{sec:ablation}, based on the experimentation on different pretraining data, we find that a small number of high-quality data with fluent texts can perform even slightly better than the large-scale noisy datasets.
We attribute this to the difference between generative pretraining and contrastive pretraining, since generative pretraining is directly learning the conditional distribution of words but not the similarity between texts and images.
Therefore, to train a high-performance multi-modal LLM, despite the quantity of data, it is crucial to prepare a high-quality dataset that satisfies: 1) it includes high-quality and fluent texts; 2) it aligns the texts and images well.

\paragraph{Tasks and prompts are crucial for zero-shot abilities.}
As shown in Section \ref{sec:ablation}, diversified prompts have a great impact on the final performance.
The essential observation behind this is that the zero-shot generality of multi-modal language models depends on the diversity of tasks involved during training.
The model can generalize to more and more unseen instructions as it sees more and more types of tasks.
This accords with the observation in text-only models \cite{radford2021learning}.

\paragraph{Balancing the correctness and language generation ability is important.}
In our experiments, we find that if the model is under-trained on downstream tasks such as VQA, it will suffer from the problem of hallucination and keep making mistakes.
While if the model is over-trained on downstream tasks, it will not be able to follow the user's instructions to generate long answers.
Therefore, it would be important to carefully balance the training data to train it so as to correctly read images and videos while keeping its generation ability.

\subsection{Limitations}

\paragraph{Evaluation}

It is hard to evaluate a multi-modal large language model since its evaluation is essentially different from traditional visual-language models.
Though we take the first step to quantitatively evaluate both the multi-modal understanding accuracy and language generation ability, it is still an open problem: \textit{how can we establish a comprehensive and automatic benchmark to evaluate existing multi-modal large language models?}

\paragraph{Training Data}
Though we have successfully collected and cleaned a mixed dataset to train our \ours, we still put a lot of effort to balance different abilities (e.g. correctness and language generation, long and short answers).
Moreover, there are still no available image-text datasets that contain long texts which are ideal for pretraining.
Besides, restricted by the computational resources that we can use, we do not conduct extensive experiments to find the optimal data combination strategy (e.g. sampling ratios, tasks, and prompts), which has been left for future work.

\paragraph{Multi-lingual}
Our model is built upon LLaMA \cite{touvron2023llama}, which is mainly trained on English corpus.
Therefore, our model is not that good at multi-lingual responses.
Though it can understand and sometimes output other languages (like shown in Figure \ref{fig:multi-lingual-demos}), it is still unexplored how to build a high-performance multi-lingual and multi-modal large language model. 

\paragraph{Safety}
Currently, we do not conduct safety checks and restrict the outputs of our model.
Therefore, the model may output contents that are not appropriate and even toxic, depending on and restricted by the data used for training.
The authors do not support the use of harmful language generation using our codes and models, like any usage on ethical, political, and racism issues.

\section{Conclusions}

In this paper, we present \ours, a multi-modal GPT4-style large language model that can take as input images/videos and responses with open-ended natural languages.
Through extensive empirical study, we show that our model outperforms other existing open-source models both in multi-modal understanding and language generation.
We also explore different factors that can affect the performance of a multi-modal large language model and conclude that: 1) for network structure, prefix-tuning is better than cross-attention to fuse different modalities; 2) instruction following is closely related to the number of tasks and prompts used for training; 3) the generative pretraining is much more sensitive the quality of training data than previous pretraining methods such as contrastive training; 4) balancing the correctness and language generation is important for multi-modal large language models.

For future work, it is promising to scale up the model to a larger size (e.g. 30B and 65B LLaMA \cite{touvron2023llama}), as well as a larger and more diversified set of instructional tasks.
Moreover, a large-scale and high-quality multi-modal dataset is also needed to train such models.
Therefore, it is worth the effort to collect such a dataset, which will be a great contribution to this area.
Multi-lingual ability and safety are also undoubtedly crucial for realistic applications.
\section*{Acknowledgements}
We would like to acknowledge Hang Li at ByteDance for his generous assistance in insightful comments in technical discussions. 
Additionally, we extend our appreciation to the colleagues at ByteDance for their efforts and support of this project. We are also thankful to the LLaMA and Vicuna teams for granting us access to their models.

\bibliographystyle{unsrtnat}
\bibliography{ref}

\begin{thebibliography}{107}
\providecommand{\natexlab}[1]{#1}
\providecommand{\url}[1]{\texttt{#1}}
\expandafter\ifx\csname urlstyle\endcsname\relax
  \providecommand{\doi}[1]{doi: #1}\else
  \providecommand{\doi}{doi: \begingroup \urlstyle{rm}\Url}\fi

\bibitem[Brown et~al.(2020)Brown, Mann, Ryder, Subbiah, Kaplan, Dhariwal,
  Neelakantan, Shyam, Sastry, Askell, et~al.]{brown2020language}
Tom Brown, Benjamin Mann, Nick Ryder, Melanie Subbiah, Jared~D Kaplan, Prafulla
  Dhariwal, Arvind Neelakantan, Pranav Shyam, Girish Sastry, Amanda Askell,
  et~al.
\newblock Language models are few-shot learners.
\newblock \emph{Advances in neural information processing systems},
  33:\penalty0 1877--1901, 2020.

\bibitem[Thoppilan et~al.(2022)Thoppilan, De~Freitas, Hall, Shazeer,
  Kulshreshtha, Cheng, Jin, Bos, Baker, Du, et~al.]{thoppilan2022lamda}
Romal Thoppilan, Daniel De~Freitas, Jamie Hall, Noam Shazeer, Apoorv
  Kulshreshtha, Heng-Tze Cheng, Alicia Jin, Taylor Bos, Leslie Baker, Yu~Du,
  et~al.
\newblock Lamda: Language models for dialog applications.
\newblock \emph{arXiv preprint arXiv:2201.08239}, 2022.

\bibitem[Hoffmann et~al.(2022{\natexlab{a}})Hoffmann, Borgeaud, Mensch,
  Buchatskaya, Cai, Rutherford, de~Las~Casas, Hendricks, Welbl, Clark,
  et~al.]{hoffmann2022empirical}
Jordan Hoffmann, Sebastian Borgeaud, Arthur Mensch, Elena Buchatskaya, Trevor
  Cai, Eliza Rutherford, Diego de~Las~Casas, Lisa~Anne Hendricks, Johannes
  Welbl, Aidan Clark, et~al.
\newblock An empirical analysis of compute-optimal large language model
  training.
\newblock \emph{Advances in Neural Information Processing Systems},
  35:\penalty0 30016--30030, 2022{\natexlab{a}}.

\bibitem[Chung et~al.(2022)Chung, Hou, Longpre, Zoph, Tay, Fedus, Li, Wang,
  Dehghani, Brahma, et~al.]{chung2022scaling}
Hyung~Won Chung, Le~Hou, Shayne Longpre, Barret Zoph, Yi~Tay, William Fedus,
  Eric Li, Xuezhi Wang, Mostafa Dehghani, Siddhartha Brahma, et~al.
\newblock Scaling instruction-finetuned language models.
\newblock \emph{arXiv preprint arXiv:2210.11416}, 2022.

\bibitem[Zhang et~al.(2022)Zhang, Roller, Goyal, Artetxe, Chen, Chen, Dewan,
  Diab, Li, Lin, et~al.]{zhang2022opt}
Susan Zhang, Stephen Roller, Naman Goyal, Mikel Artetxe, Moya Chen, Shuohui
  Chen, Christopher Dewan, Mona Diab, Xian Li, Xi~Victoria Lin, et~al.
\newblock Opt: Open pre-trained transformer language models.
\newblock \emph{arXiv preprint arXiv:2205.01068}, 2022.

\bibitem[Chowdhery et~al.(2022)Chowdhery, Narang, Devlin, Bosma, Mishra,
  Roberts, Barham, Chung, Sutton, Gehrmann, et~al.]{chowdhery2022palm}
Aakanksha Chowdhery, Sharan Narang, Jacob Devlin, Maarten Bosma, Gaurav Mishra,
  Adam Roberts, Paul Barham, Hyung~Won Chung, Charles Sutton, Sebastian
  Gehrmann, et~al.
\newblock Palm: Scaling language modeling with pathways.
\newblock \emph{arXiv preprint arXiv:2204.02311}, 2022.

\bibitem[Ouyang et~al.(2022)Ouyang, Wu, Jiang, Almeida, Wainwright, Mishkin,
  Zhang, Agarwal, Slama, Ray, et~al.]{ouyang2022training}
Long Ouyang, Jeffrey Wu, Xu~Jiang, Diogo Almeida, Carroll Wainwright, Pamela
  Mishkin, Chong Zhang, Sandhini Agarwal, Katarina Slama, Alex Ray, et~al.
\newblock Training language models to follow instructions with human feedback.
\newblock \emph{Advances in Neural Information Processing Systems},
  35:\penalty0 27730--27744, 2022.

\bibitem[Scao et~al.(2022)Scao, Fan, Akiki, Pavlick, Ili{\'c}, Hesslow,
  Castagn{\'e}, Luccioni, Yvon, Gall{\'e}, et~al.]{scao2022bloom}
Teven~Le Scao, Angela Fan, Christopher Akiki, Ellie Pavlick, Suzana Ili{\'c},
  Daniel Hesslow, Roman Castagn{\'e}, Alexandra~Sasha Luccioni, Fran{\c{c}}ois
  Yvon, Matthias Gall{\'e}, et~al.
\newblock Bloom: A 176b-parameter open-access multilingual language model.
\newblock \emph{arXiv preprint arXiv:2211.05100}, 2022.

\bibitem[Du et~al.(2022)Du, Qian, Liu, Ding, Qiu, Yang, and Tang]{du2022glm}
Zhengxiao Du, Yujie Qian, Xiao Liu, Ming Ding, Jiezhong Qiu, Zhilin Yang, and
  Jie Tang.
\newblock Glm: General language model pretraining with autoregressive blank
  infilling.
\newblock In \emph{Proceedings of the 60th Annual Meeting of the Association
  for Computational Linguistics (Volume 1: Long Papers)}, pages 320--335, 2022.

\bibitem[Zeng et~al.(2022{\natexlab{a}})Zeng, Liu, Du, Wang, Lai, Ding, Yang,
  Xu, Zheng, Xia, et~al.]{zeng2022glm}
Aohan Zeng, Xiao Liu, Zhengxiao Du, Zihan Wang, Hanyu Lai, Ming Ding, Zhuoyi
  Yang, Yifan Xu, Wendi Zheng, Xiao Xia, et~al.
\newblock Glm-130b: An open bilingual pre-trained model.
\newblock \emph{arXiv preprint arXiv:2210.02414}, 2022{\natexlab{a}}.

\bibitem[Iyer et~al.(2022)Iyer, Lin, Pasunuru, Mihaylov, Simig, Yu, Shuster,
  Wang, Liu, Koura, et~al.]{iyer2022opt}
Srinivasan Iyer, Xi~Victoria Lin, Ramakanth Pasunuru, Todor Mihaylov,
  D{\'a}niel Simig, Ping Yu, Kurt Shuster, Tianlu Wang, Qing Liu, Punit~Singh
  Koura, et~al.
\newblock Opt-iml: Scaling language model instruction meta learning through the
  lens of generalization.
\newblock \emph{arXiv preprint arXiv:2212.12017}, 2022.

\bibitem[Touvron et~al.(2023)Touvron, Lavril, Izacard, Martinet, Lachaux,
  Lacroix, Rozi{\`e}re, Goyal, Hambro, Azhar, et~al.]{touvron2023llama}
Hugo Touvron, Thibaut Lavril, Gautier Izacard, Xavier Martinet, Marie-Anne
  Lachaux, Timoth{\'e}e Lacroix, Baptiste Rozi{\`e}re, Naman Goyal, Eric
  Hambro, Faisal Azhar, et~al.
\newblock Llama: Open and efficient foundation language models.
\newblock \emph{arXiv preprint arXiv:2302.13971}, 2023.

\bibitem[Anil et~al.(2023)Anil, Dai, Firat, Johnson, Lepikhin, Passos, Shakeri,
  Taropa, Bailey, Chen, et~al.]{anil2023palm}
Rohan Anil, Andrew~M Dai, Orhan Firat, Melvin Johnson, Dmitry Lepikhin,
  Alexandre Passos, Siamak Shakeri, Emanuel Taropa, Paige Bailey, Zhifeng Chen,
  et~al.
\newblock Palm 2 technical report.
\newblock \emph{arXiv preprint arXiv:2305.10403}, 2023.

\bibitem[Wang et~al.(2022)Wang, Kordi, Mishra, Liu, Smith, Khashabi, and
  Hajishirzi]{wang2022self}
Yizhong Wang, Yeganeh Kordi, Swaroop Mishra, Alisa Liu, Noah~A Smith, Daniel
  Khashabi, and Hannaneh Hajishirzi.
\newblock Self-instruct: Aligning language model with self generated
  instructions.
\newblock \emph{arXiv preprint arXiv:2212.10560}, 2022.

\bibitem[Chiang et~al.(2023)Chiang, Li, Lin, Sheng, Wu, Zhang, Zheng, Zhuang,
  Zhuang, Gonzalez, Stoica, and Xing]{vicuna2023}
Wei-Lin Chiang, Zhuohan Li, Zi~Lin, Ying Sheng, Zhanghao Wu, Hao Zhang, Lianmin
  Zheng, Siyuan Zhuang, Yonghao Zhuang, Joseph~E. Gonzalez, Ion Stoica, and
  Eric~P. Xing.
\newblock Vicuna: An open-source chatbot impressing gpt-4 with 90\%* chatgpt
  quality, March 2023.
\newblock URL \url{https://lmsys.org/blog/2023-03-30-vicuna/}.

\bibitem[Xu et~al.(2023)Xu, Guo, Duan, and McAuley]{xu2023baize}
Canwen Xu, Daya Guo, Nan Duan, and Julian McAuley.
\newblock Baize: An open-source chat model with parameter-efficient tuning on
  self-chat data.
\newblock \emph{arXiv preprint arXiv:2304.01196}, 2023.

\bibitem[Peng et~al.(2023)Peng, Li, He, Galley, and Gao]{peng2023instruction}
Baolin Peng, Chunyuan Li, Pengcheng He, Michel Galley, and Jianfeng Gao.
\newblock Instruction tuning with gpt-4.
\newblock \emph{arXiv preprint arXiv:2304.03277}, 2023.

\bibitem[Li et~al.(2023{\natexlab{a}})Li, Li, Savarese, and Hoi]{li2023blip}
Junnan Li, Dongxu Li, Silvio Savarese, and Steven Hoi.
\newblock Blip-2: Bootstrapping language-image pre-training with frozen image
  encoders and large language models.
\newblock \emph{arXiv preprint arXiv:2301.12597}, 2023{\natexlab{a}}.

\bibitem[Alayrac et~al.(2022)Alayrac, Donahue, Luc, Miech, Barr, Hasson, Lenc,
  Mensch, Millican, Reynolds, et~al.]{alayrac2022flamingo}
Jean-Baptiste Alayrac, Jeff Donahue, Pauline Luc, Antoine Miech, Iain Barr,
  Yana Hasson, Karel Lenc, Arthur Mensch, Katherine Millican, Malcolm Reynolds,
  et~al.
\newblock Flamingo: a visual language model for few-shot learning.
\newblock \emph{Advances in Neural Information Processing Systems},
  35:\penalty0 23716--23736, 2022.

\bibitem[Gong et~al.(2023)Gong, Lyu, Zhang, Wang, Zheng, Zhao, Liu, Zhang, Luo,
  and Chen]{gong2023multimodal}
Tao Gong, Chengqi Lyu, Shilong Zhang, Yudong Wang, Miao Zheng, Qian Zhao,
  Kuikun Liu, Wenwei Zhang, Ping Luo, and Kai Chen.
\newblock Multimodal-gpt: A vision and language model for dialogue with humans.
\newblock \emph{arXiv preprint arXiv:2305.04790}, 2023.

\bibitem[Li et~al.(2023{\natexlab{b}})Li, Zhang, Chen, Wang, Yang, and
  Liu]{li2023otter}
Bo~Li, Yuanhan Zhang, Liangyu Chen, Jinghao Wang, Jingkang Yang, and Ziwei Liu.
\newblock Otter: A multi-modal model with in-context instruction tuning.
\newblock \emph{arXiv preprint arXiv:2305.03726}, 2023{\natexlab{b}}.

\bibitem[Ye et~al.(2023)Ye, Xu, Xu, Ye, Yan, Zhou, Wang, Hu, Shi, Shi,
  et~al.]{ye2023mplug}
Qinghao Ye, Haiyang Xu, Guohai Xu, Jiabo Ye, Ming Yan, Yiyang Zhou, Junyang
  Wang, Anwen Hu, Pengcheng Shi, Yaya Shi, et~al.
\newblock mplug-owl: Modularization empowers large language models with
  multimodality.
\newblock \emph{arXiv preprint arXiv:2304.14178}, 2023.

\bibitem[Zhu et~al.(2023)Zhu, Chen, Shen, Li, and Elhoseiny]{zhu2023minigpt}
Deyao Zhu, Jun Chen, Xiaoqian Shen, Xiang Li, and Mohamed Elhoseiny.
\newblock Minigpt-4: Enhancing vision-language understanding with advanced
  large language models.
\newblock \emph{arXiv preprint arXiv:2304.10592}, 2023.

\bibitem[Dai et~al.(2023)Dai, Li, Li, Tiong, Zhao, Wang, Li, Fung, and
  Hoi]{dai2023instructblip}
Wenliang Dai, Junnan Li, Dongxu Li, Anthony Meng~Huat Tiong, Junqi Zhao,
  Weisheng Wang, Boyang Li, Pascale Fung, and Steven Hoi.
\newblock Instructblip: Towards general-purpose vision-language models with
  instruction tuning.
\newblock \emph{arXiv preprint arXiv:2305.06500}, 2023.

\bibitem[Chen et~al.(2023)Chen, Zheng, Wang, Xu, Huang, Pan, Wang, Wang, Qiao,
  Lu, et~al.]{chen2023videollm}
Guo Chen, Yin-Dong Zheng, Jiahao Wang, Jilan Xu, Yifei Huang, Junting Pan,
  Yi~Wang, Yali Wang, Yu~Qiao, Tong Lu, et~al.
\newblock Videollm: Modeling video sequence with large language models.
\newblock \emph{arXiv preprint arXiv:2305.13292}, 2023.

\bibitem[Maaz et~al.(2023)Maaz, Rasheed, Khan, and Khan]{maaz2023video}
Muhammad Maaz, Hanoona Rasheed, Salman Khan, and Fahad~Shahbaz Khan.
\newblock Video-chatgpt: Towards detailed video understanding via large vision
  and language models.
\newblock \emph{arXiv preprint arXiv:2306.05424}, 2023.

\bibitem[Zhao et~al.(2023)Zhao, Misra, Kr{\"a}henb{\"u}hl, and
  Girdhar]{zhao2023learning}
Yue Zhao, Ishan Misra, Philipp Kr{\"a}henb{\"u}hl, and Rohit Girdhar.
\newblock Learning video representations from large language models.
\newblock In \emph{Proceedings of the IEEE/CVF Conference on Computer Vision
  and Pattern Recognition}, pages 6586--6597, 2023.

\bibitem[Luo et~al.(2023)Luo, Zhao, Yang, Dong, Qiu, Lu, Wang, and
  Wei]{luo2023valley}
Ruipu Luo, Ziwang Zhao, Min Yang, Junwei Dong, Minghui Qiu, Pengcheng Lu, Tao
  Wang, and Zhongyu Wei.
\newblock Valley: Video assistant with large language model enhanced ability.
\newblock \emph{arXiv preprint arXiv:2306.07207}, 2023.

\bibitem[Zhang et~al.(2023{\natexlab{a}})Zhang, Li, and Bing]{zhang2023video}
Hang Zhang, Xin Li, and Lidong Bing.
\newblock Video-llama: An instruction-tuned audio-visual language model for
  video understanding.
\newblock \emph{arXiv preprint arXiv:2306.02858}, 2023{\natexlab{a}}.

\bibitem[Huang et~al.(2023{\natexlab{a}})Huang, Li, Yang, Shi, Chang, Ye, Wu,
  Hong, Huang, Liu, et~al.]{huang2023audiogpt}
Rongjie Huang, Mingze Li, Dongchao Yang, Jiatong Shi, Xuankai Chang, Zhenhui
  Ye, Yuning Wu, Zhiqing Hong, Jiawei Huang, Jinglin Liu, et~al.
\newblock Audiogpt: Understanding and generating speech, music, sound, and
  talking head.
\newblock \emph{arXiv preprint arXiv:2304.12995}, 2023{\natexlab{a}}.

\bibitem[Wang et~al.(2023)Wang, Chen, Chen, Wu, Zhu, Zeng, Luo, Lu, Zhou, Qiao,
  et~al.]{wang2023visionllm}
Wenhai Wang, Zhe Chen, Xiaokang Chen, Jiannan Wu, Xizhou Zhu, Gang Zeng, Ping
  Luo, Tong Lu, Jie Zhou, Yu~Qiao, et~al.
\newblock Visionllm: Large language model is also an open-ended decoder for
  vision-centric tasks.
\newblock \emph{arXiv preprint arXiv:2305.11175}, 2023.

\bibitem[Jin et~al.(2023)Jin, Tan, Yang, Liu, Song, Wang, and
  Fu]{jin2023alphablock}
Chuhao Jin, Wenhui Tan, Jiange Yang, Bei Liu, Ruihua Song, Limin Wang, and
  Jianlong Fu.
\newblock Alphablock: Embodied finetuning for vision-language reasoning in
  robot manipulation.
\newblock \emph{arXiv preprint arXiv:2305.18898}, 2023.

\bibitem[Driess et~al.(2023)Driess, Xia, Sajjadi, Lynch, Chowdhery, Ichter,
  Wahid, Tompson, Vuong, Yu, et~al.]{driess2023palm}
Danny Driess, Fei Xia, Mehdi~SM Sajjadi, Corey Lynch, Aakanksha Chowdhery,
  Brian Ichter, Ayzaan Wahid, Jonathan Tompson, Quan Vuong, Tianhe Yu, et~al.
\newblock Palm-e: An embodied multimodal language model.
\newblock \emph{arXiv preprint arXiv:2303.03378}, 2023.

\bibitem[Jiang et~al.(2022)Jiang, Gupta, Zhang, Wang, Dou, Chen, Fei-Fei,
  Anandkumar, Zhu, and Fan]{jiang2022vima}
Yunfan Jiang, Agrim Gupta, Zichen Zhang, Guanzhi Wang, Yongqiang Dou, Yanjun
  Chen, Li~Fei-Fei, Anima Anandkumar, Yuke Zhu, and Linxi Fan.
\newblock Vima: General robot manipulation with multimodal prompts.
\newblock \emph{arXiv preprint arXiv:2210.03094}, 2022.

\bibitem[Li et~al.(2021)Li, Selvaraju, Gotmare, Joty, Xiong, and
  Hoi]{li2021align}
Junnan Li, Ramprasaath Selvaraju, Akhilesh Gotmare, Shafiq Joty, Caiming Xiong,
  and Steven Chu~Hong Hoi.
\newblock Align before fuse: Vision and language representation learning with
  momentum distillation.
\newblock \emph{Advances in neural information processing systems},
  34:\penalty0 9694--9705, 2021.

\bibitem[Bao et~al.(2022)Bao, Wang, Dong, Liu, Mohammed, Aggarwal, Som, Piao,
  and Wei]{bao2022vlmo}
Hangbo Bao, Wenhui Wang, Li~Dong, Qiang Liu, Owais~Khan Mohammed, Kriti
  Aggarwal, Subhojit Som, Songhao Piao, and Furu Wei.
\newblock Vlmo: Unified vision-language pre-training with
  mixture-of-modality-experts.
\newblock \emph{Advances in Neural Information Processing Systems},
  35:\penalty0 32897--32912, 2022.

\bibitem[Zeng et~al.(2021)Zeng, Zhang, and Li]{zeng2021multi}
Yan Zeng, Xinsong Zhang, and Hang Li.
\newblock Multi-grained vision language pre-training: Aligning texts with
  visual concepts.
\newblock \emph{arXiv preprint arXiv:2111.08276}, 2021.

\bibitem[Zeng et~al.(2022{\natexlab{b}})Zeng, Zhang, Li, Wang, Zhang, and
  Zhou]{zeng2022x}
Yan Zeng, Xinsong Zhang, Hang Li, Jiawei Wang, Jipeng Zhang, and Wangchunshu
  Zhou.
\newblock X$^2$-vlm: All-in-one pre-trained model for vision-language tasks.
\newblock \emph{arXiv preprint arXiv:2211.12402}, 2022{\natexlab{b}}.

\bibitem[Li et~al.(2022)Li, Li, Xiong, and Hoi]{li2022blip}
Junnan Li, Dongxu Li, Caiming Xiong, and Steven Hoi.
\newblock Blip: Bootstrapping language-image pre-training for unified
  vision-language understanding and generation.
\newblock In \emph{International Conference on Machine Learning}, pages
  12888--12900. PMLR, 2022.

\bibitem[Byeon et~al.(2022)Byeon, Park, Kim, Lee, Baek, and
  Kim]{kakaobrain2022coyo}
Minwoo Byeon, Beomhee Park, Haecheon Kim, Sungjun Lee, Woonhyuk Baek, and
  Saehoon Kim.
\newblock Coyo-700m: Image-text pair dataset.
\newblock \url{https://github.com/kakaobrain/coyo-dataset}, 2022.

\bibitem[Gadre et~al.(2023)Gadre, Ilharco, Fang, Hayase, Smyrnis, Nguyen,
  Marten, Wortsman, Ghosh, Zhang, et~al.]{gadre2023datacomp}
Samir~Yitzhak Gadre, Gabriel Ilharco, Alex Fang, Jonathan Hayase, Georgios
  Smyrnis, Thao Nguyen, Ryan Marten, Mitchell Wortsman, Dhruba Ghosh, Jieyu
  Zhang, et~al.
\newblock Datacomp: In search of the next generation of multimodal datasets.
\newblock \emph{arXiv preprint arXiv:2304.14108}, 2023.

\bibitem[Antol et~al.(2015)Antol, Agrawal, Lu, Mitchell, Batra, Zitnick, and
  Parikh]{antol2015vqa}
Stanislaw Antol, Aishwarya Agrawal, Jiasen Lu, Margaret Mitchell, Dhruv Batra,
  C~Lawrence Zitnick, and Devi Parikh.
\newblock Vqa: Visual question answering.
\newblock In \emph{Proceedings of the IEEE international conference on computer
  vision}, pages 2425--2433, 2015.

\bibitem[Zhang et~al.(2016)Zhang, Goyal, Summers-Stay, Batra, and
  Parikh]{zhang2016yin}
Peng Zhang, Yash Goyal, Douglas Summers-Stay, Dhruv Batra, and Devi Parikh.
\newblock Yin and yang: Balancing and answering binary visual questions.
\newblock In \emph{Proceedings of the IEEE conference on computer vision and
  pattern recognition}, pages 5014--5022, 2016.

\bibitem[Radford et~al.(2019)Radford, Wu, Child, Luan, Amodei, Sutskever,
  et~al.]{radford2019language}
Alec Radford, Jeffrey Wu, Rewon Child, David Luan, Dario Amodei, Ilya
  Sutskever, et~al.
\newblock Language models are unsupervised multitask learners.
\newblock \emph{OpenAI blog}, 1\penalty0 (8):\penalty0 9, 2019.

\bibitem[OpenAI(2023)]{openai2023gpt}
OpenAI.
\newblock Gpt-4 technical report.
\newblock \emph{arXiv}, page 2303.08774, 2023.

\bibitem[Chen et~al.(2020)Chen, Kornblith, Norouzi, and Hinton]{chen2020simple}
Ting Chen, Simon Kornblith, Mohammad Norouzi, and Geoffrey Hinton.
\newblock A simple framework for contrastive learning of visual
  representations.
\newblock In \emph{International conference on machine learning}, pages
  1597--1607. PMLR, 2020.

\bibitem[Radford et~al.(2021)Radford, Kim, Hallacy, Ramesh, Goh, Agarwal,
  Sastry, Askell, Mishkin, Clark, et~al.]{radford2021learning}
Alec Radford, Jong~Wook Kim, Chris Hallacy, Aditya Ramesh, Gabriel Goh,
  Sandhini Agarwal, Girish Sastry, Amanda Askell, Pamela Mishkin, Jack Clark,
  et~al.
\newblock Learning transferable visual models from natural language
  supervision.
\newblock In \emph{International conference on machine learning}, pages
  8748--8763. PMLR, 2021.

\bibitem[Ba et~al.(2016)Ba, Kiros, and Hinton]{ba2016layer}
Jimmy~Lei Ba, Jamie~Ryan Kiros, and Geoffrey~E Hinton.
\newblock Layer normalization.
\newblock \emph{arXiv preprint arXiv:1607.06450}, 2016.

\bibitem[Elfwing et~al.(2018)Elfwing, Uchibe, and Doya]{elfwing2018sigmoid}
Stefan Elfwing, Eiji Uchibe, and Kenji Doya.
\newblock Sigmoid-weighted linear units for neural network function
  approximation in reinforcement learning.
\newblock \emph{Neural Networks}, 107:\penalty0 3--11, 2018.

\bibitem[Fang et~al.(2023)Fang, Wang, Xie, Sun, Wu, Wang, Huang, Wang, and
  Cao]{fang2023eva}
Yuxin Fang, Wen Wang, Binhui Xie, Quan Sun, Ledell Wu, Xinggang Wang, Tiejun
  Huang, Xinlong Wang, and Yue Cao.
\newblock Eva: Exploring the limits of masked visual representation learning at
  scale.
\newblock In \emph{Proceedings of the IEEE/CVF Conference on Computer Vision
  and Pattern Recognition}, pages 19358--19369, 2023.

\bibitem[Sun et~al.(2023)Sun, Fang, Wu, Wang, and Cao]{sun2023eva}
Quan Sun, Yuxin Fang, Ledell Wu, Xinlong Wang, and Yue Cao.
\newblock Eva-clip: Improved training techniques for clip at scale.
\newblock \emph{arXiv preprint arXiv:2303.15389}, 2023.

\bibitem[Jaegle et~al.(2021)Jaegle, Gimeno, Brock, Vinyals, Zisserman, and
  Carreira]{jaegle2021perceiver}
Andrew Jaegle, Felix Gimeno, Andy Brock, Oriol Vinyals, Andrew Zisserman, and
  Joao Carreira.
\newblock Perceiver: General perception with iterative attention.
\newblock In \emph{International conference on machine learning}, pages
  4651--4664. PMLR, 2021.

\bibitem[Changpinyo et~al.(2021)Changpinyo, Sharma, Ding, and
  Soricut]{changpinyo2021conceptual}
Soravit Changpinyo, Piyush Sharma, Nan Ding, and Radu Soricut.
\newblock Conceptual 12m: Pushing web-scale image-text pre-training to
  recognize long-tail visual concepts.
\newblock In \emph{Proceedings of the IEEE/CVF Conference on Computer Vision
  and Pattern Recognition}, pages 3558--3568, 2021.

\bibitem[Sharma et~al.(2018)Sharma, Ding, Goodman, and
  Soricut]{sharma2018conceptual}
Piyush Sharma, Nan Ding, Sebastian Goodman, and Radu Soricut.
\newblock Conceptual captions: A cleaned, hypernymed, image alt-text dataset
  for automatic image captioning.
\newblock In \emph{Proceedings of the 56th Annual Meeting of the Association
  for Computational Linguistics (Volume 1: Long Papers)}, pages 2556--2565,
  2018.

\bibitem[Ordonez et~al.(2011)Ordonez, Kulkarni, and Berg]{ordonez2011im2text}
Vicente Ordonez, Girish Kulkarni, and Tamara Berg.
\newblock Im2text: Describing images using 1 million captioned photographs.
\newblock \emph{Advances in neural information processing systems}, 24, 2011.

\bibitem[Liu et~al.(2023{\natexlab{a}})Liu, Li, Wu, and Lee]{liu2023llava}
Haotian Liu, Chunyuan Li, Qingyang Wu, and Yong~Jae Lee.
\newblock Visual instruction tuning.
\newblock \emph{arXiv preprint arXiv:2304.08485}, 2023{\natexlab{a}}.

\bibitem[Chen et~al.(2022)Chen, Wang, Changpinyo, Piergiovanni, Padlewski,
  Salz, Goodman, Grycner, Mustafa, Beyer, et~al.]{chen2022pali}
Xi~Chen, Xiao Wang, Soravit Changpinyo, AJ~Piergiovanni, Piotr Padlewski,
  Daniel Salz, Sebastian Goodman, Adam Grycner, Basil Mustafa, Lucas Beyer,
  et~al.
\newblock Pali: A jointly-scaled multilingual language-image model.
\newblock \emph{arXiv preprint arXiv:2209.06794}, 2022.

\bibitem[Chen et~al.(2015)Chen, Fang, Lin, Vedantam, Gupta, Doll{\'a}r, and
  Zitnick]{chen2015microsoft}
Xinlei Chen, Hao Fang, Tsung-Yi Lin, Ramakrishna Vedantam, Saurabh Gupta, Piotr
  Doll{\'a}r, and C~Lawrence Zitnick.
\newblock Microsoft coco captions: Data collection and evaluation server.
\newblock \emph{arXiv preprint arXiv:1504.00325}, 2015.

\bibitem[Kafle and Kanan(2017)]{kafle2017analysis}
Kushal Kafle and Christopher Kanan.
\newblock An analysis of visual question answering algorithms.
\newblock In \emph{ICCV}, 2017.

\bibitem[Mishra et~al.(2019)Mishra, Shekhar, Singh, and
  Chakraborty]{mishra2019ocr}
Anand Mishra, Shashank Shekhar, Ajeet~Kumar Singh, and Anirban Chakraborty.
\newblock Ocr-vqa: Visual question answering by reading text in images.
\newblock In \emph{2019 international conference on document analysis and
  recognition (ICDAR)}, pages 947--952. IEEE, 2019.

\bibitem[Zhou et~al.(2017)Zhou, Lapedriza, Khosla, Oliva, and
  Torralba]{zhou2017places}
Bolei Zhou, Agata Lapedriza, Aditya Khosla, Aude Oliva, and Antonio Torralba.
\newblock Places: A 10 million image database for scene recognition.
\newblock \emph{IEEE transactions on pattern analysis and machine
  intelligence}, 40\penalty0 (6):\penalty0 1452--1464, 2017.

\bibitem[Chen and Dolan(2011)]{chen2011collecting}
David Chen and William~B Dolan.
\newblock Collecting highly parallel data for paraphrase evaluation.
\newblock In \emph{Proceedings of the 49th annual meeting of the association
  for computational linguistics: human language technologies}, pages 190--200,
  2011.

\bibitem[Xu et~al.(2016)Xu, Mei, Yao, and Rui]{xu2016msr}
Jun Xu, Tao Mei, Ting Yao, and Yong Rui.
\newblock Msr-vtt: A large video description dataset for bridging video and
  language.
\newblock In \emph{Proceedings of the IEEE conference on computer vision and
  pattern recognition}, pages 5288--5296, 2016.

\bibitem[Goyal et~al.(2017)Goyal, Ebrahimi~Kahou, Michalski, Materzynska,
  Westphal, Kim, Haenel, Fruend, Yianilos, Mueller-Freitag,
  et~al.]{goyal2017something}
Raghav Goyal, Samira Ebrahimi~Kahou, Vincent Michalski, Joanna Materzynska,
  Susanne Westphal, Heuna Kim, Valentin Haenel, Ingo Fruend, Peter Yianilos,
  Moritz Mueller-Freitag, et~al.
\newblock The" something something" video database for learning and evaluating
  visual common sense.
\newblock In \emph{Proceedings of the IEEE international conference on computer
  vision}, pages 5842--5850, 2017.

\bibitem[Yang* et~al.(2023)Yang*, Li*, Wang*, Lin*, Azarnasab*, Ahmed*, Liu,
  Liu, Zeng, and Wang]{yang2023mmreact}
Zhengyuan Yang*, Linjie Li*, Jianfeng Wang*, Kevin Lin*, Ehsan Azarnasab*,
  Faisal Ahmed*, Zicheng Liu, Ce~Liu, Michael Zeng, and Lijuan Wang.
\newblock Mm-react: Prompting chatgpt for multimodal reasoning and action.
\newblock 2023.

\bibitem[Fu et~al.(2023)Fu, Chen, Shen, Qin, Zhang, Lin, Qiu, Lin, Yang, Zheng,
  Li, Sun, and Ji]{fu2023mme}
Chaoyou Fu, Peixian Chen, Yunhang Shen, Yulei Qin, Mengdan Zhang, Xu~Lin,
  Zhenyu Qiu, Wei Lin, Jinrui Yang, Xiawu Zheng, Ke~Li, Xing Sun, and Rongrong
  Ji.
\newblock Mme: A comprehensive evaluation benchmark for multimodal large
  language models.
\newblock \emph{arXiv preprint arXiv:2306.13394}, 2023.

\bibitem[Awadalla et~al.(2023)Awadalla, Gao, Gardner, Hessel, Hanafy, Zhu,
  Marathe, Bitton, Gadre, Jitsev, Kornblith, Koh, Ilharco, Wortsman, and
  Schmidt]{anas_awadalla_2023_7733589}
Anas Awadalla, Irena Gao, Joshua Gardner, Jack Hessel, Yusuf Hanafy, Wanrong
  Zhu, Kalyani Marathe, Yonatan Bitton, Samir Gadre, Jenia Jitsev, Simon
  Kornblith, Pang~Wei Koh, Gabriel Ilharco, Mitchell Wortsman, and Ludwig
  Schmidt.
\newblock Openflamingo, March 2023.
\newblock URL \url{https://doi.org/10.5281/zenodo.7733589}.

\bibitem[Hoffmann et~al.(2022{\natexlab{b}})Hoffmann, Borgeaud, Mensch,
  Buchatskaya, Cai, Rutherford, Casas, Hendricks, Welbl, Clark,
  et~al.]{hoffmann2022training}
Jordan Hoffmann, Sebastian Borgeaud, Arthur Mensch, Elena Buchatskaya, Trevor
  Cai, Eliza Rutherford, Diego de~Las Casas, Lisa~Anne Hendricks, Johannes
  Welbl, Aidan Clark, et~al.
\newblock Training compute-optimal large language models.
\newblock \emph{arXiv preprint arXiv:2203.15556}, 2022{\natexlab{b}}.

\bibitem[Raffel et~al.(2020)Raffel, Shazeer, Roberts, Lee, Narang, Matena,
  Zhou, Li, and Liu]{raffel2020exploring}
Colin Raffel, Noam Shazeer, Adam Roberts, Katherine Lee, Sharan Narang, Michael
  Matena, Yanqi Zhou, Wei Li, and Peter~J Liu.
\newblock Exploring the limits of transfer learning with a unified text-to-text
  transformer.
\newblock \emph{The Journal of Machine Learning Research}, 21\penalty0
  (1):\penalty0 5485--5551, 2020.

\bibitem[Kaplan et~al.(2020)Kaplan, McCandlish, Henighan, Brown, Chess, Child,
  Gray, Radford, Wu, and Amodei]{kaplan2020scaling}
Jared Kaplan, Sam McCandlish, Tom Henighan, Tom~B Brown, Benjamin Chess, Rewon
  Child, Scott Gray, Alec Radford, Jeffrey Wu, and Dario Amodei.
\newblock Scaling laws for neural language models.
\newblock \emph{arXiv preprint arXiv:2001.08361}, 2020.

\bibitem[Wu et~al.(2023)Wu, Yin, Qi, Wang, Tang, and Duan]{wu2023visual}
Chenfei Wu, Shengming Yin, Weizhen Qi, Xiaodong Wang, Zecheng Tang, and Nan
  Duan.
\newblock Visual chatgpt: Talking, drawing and editing with visual foundation
  models.
\newblock \emph{arXiv preprint arXiv:2303.04671}, 2023.

\bibitem[Yang et~al.(2023)Yang, Li, Wang, Lin, Azarnasab, Ahmed, Liu, Liu,
  Zeng, and Wang]{yang2023mm}
Zhengyuan Yang, Linjie Li, Jianfeng Wang, Kevin Lin, Ehsan Azarnasab, Faisal
  Ahmed, Zicheng Liu, Ce~Liu, Michael Zeng, and Lijuan Wang.
\newblock Mm-react: Prompting chatgpt for multimodal reasoning and action.
\newblock \emph{arXiv preprint arXiv:2303.11381}, 2023.

\bibitem[Shen et~al.(2023)Shen, Song, Tan, Li, Lu, and
  Zhuang]{shen2023hugginggpt}
Yongliang Shen, Kaitao Song, Xu~Tan, Dongsheng Li, Weiming Lu, and Yueting
  Zhuang.
\newblock Hugginggpt: Solving ai tasks with chatgpt and its friends in
  huggingface.
\newblock \emph{arXiv preprint arXiv:2303.17580}, 2023.

\bibitem[Liu et~al.(2023{\natexlab{b}})Liu, He, Wang, Wang, Wang, Chen, Zhang,
  Yang, Li, Yu, et~al.]{liu2023internchat}
Zhaoyang Liu, Yinan He, Wenhai Wang, Weiyun Wang, Yi~Wang, Shoufa Chen,
  Qinglong Zhang, Yang Yang, Qingyun Li, Jiashuo Yu, et~al.
\newblock Internchat: Solving vision-centric tasks by interacting with chatbots
  beyond language.
\newblock \emph{arXiv preprint arXiv:2305.05662}, 2023{\natexlab{b}}.

\bibitem[Ahn et~al.(2022)Ahn, Brohan, Brown, Chebotar, Cortes, David, Finn, Fu,
  Gopalakrishnan, Hausman, et~al.]{ahn2022can}
Michael Ahn, Anthony Brohan, Noah Brown, Yevgen Chebotar, Omar Cortes, Byron
  David, Chelsea Finn, Chuyuan Fu, Keerthana Gopalakrishnan, Karol Hausman,
  et~al.
\newblock Do as i can, not as i say: Grounding language in robotic affordances.
\newblock \emph{arXiv preprint arXiv:2204.01691}, 2022.

\bibitem[Huang et~al.(2022)Huang, Xia, Xiao, Chan, Liang, Florence, Zeng,
  Tompson, Mordatch, Chebotar, et~al.]{huang2022inner}
Wenlong Huang, Fei Xia, Ted Xiao, Harris Chan, Jacky Liang, Pete Florence, Andy
  Zeng, Jonathan Tompson, Igor Mordatch, Yevgen Chebotar, et~al.
\newblock Inner monologue: Embodied reasoning through planning with language
  models.
\newblock \emph{arXiv preprint arXiv:2207.05608}, 2022.

\bibitem[Brock et~al.(2021)Brock, De, Smith, and Simonyan]{brock2021high}
Andy Brock, Soham De, Samuel~L Smith, and Karen Simonyan.
\newblock High-performance large-scale image recognition without normalization.
\newblock In \emph{International Conference on Machine Learning}, pages
  1059--1071. PMLR, 2021.

\bibitem[Huang et~al.(2023{\natexlab{b}})Huang, Dong, Wang, Hao, Singhal, Ma,
  Lv, Cui, Mohammed, Liu, et~al.]{huang2023language}
Shaohan Huang, Li~Dong, Wenhui Wang, Yaru Hao, Saksham Singhal, Shuming Ma,
  Tengchao Lv, Lei Cui, Owais~Khan Mohammed, Qiang Liu, et~al.
\newblock Language is not all you need: Aligning perception with language
  models.
\newblock \emph{arXiv preprint arXiv:2302.14045}, 2023{\natexlab{b}}.

\bibitem[Schuhmann et~al.(2022)Schuhmann, Beaumont, Vencu, Gordon, Wightman,
  Cherti, Coombes, Katta, Mullis, Wortsman, et~al.]{schuhmann2022laion}
Christoph Schuhmann, Romain Beaumont, Richard Vencu, Cade Gordon, Ross
  Wightman, Mehdi Cherti, Theo Coombes, Aarush Katta, Clayton Mullis, Mitchell
  Wortsman, et~al.
\newblock Laion-5b: An open large-scale dataset for training next generation
  image-text models.
\newblock \emph{arXiv preprint arXiv:2210.08402}, 2022.

\bibitem[Gao et~al.(2020)Gao, Biderman, Black, Golding, Hoppe, Foster, Phang,
  He, Thite, Nabeshima, Presser, and Leahy]{pile}
Leo Gao, Stella Biderman, Sid Black, Laurence Golding, Travis Hoppe, Charles
  Foster, Jason Phang, Horace He, Anish Thite, Noa Nabeshima, Shawn Presser,
  and Connor Leahy.
\newblock The {P}ile: An 800gb dataset of diverse text for language modeling.
\newblock \emph{arXiv preprint arXiv:2101.00027}, 2020.

\bibitem[Rasley et~al.(2020)Rasley, Rajbhandari, Ruwase, and
  He]{rasley2020deepspeed}
Jeff Rasley, Samyam Rajbhandari, Olatunji Ruwase, and Yuxiong He.
\newblock Deepspeed: System optimizations enable training deep learning models
  with over 100 billion parameters.
\newblock In \emph{Proceedings of the 26th ACM SIGKDD International Conference
  on Knowledge Discovery \& Data Mining}, pages 3505--3506, 2020.

\bibitem[Gao et~al.(2023)Gao, Han, Zhang, Lin, Geng, Zhou, Zhang, Lu, He, Yue,
  et~al.]{gao2023llama}
Peng Gao, Jiaming Han, Renrui Zhang, Ziyi Lin, Shijie Geng, Aojun Zhou, Wei
  Zhang, Pan Lu, Conghui He, Xiangyu Yue, et~al.
\newblock Llama-adapter v2: Parameter-efficient visual instruction model.
\newblock \emph{arXiv preprint arXiv:2304.15010}, 2023.

\bibitem[Sidorov et~al.(2020)Sidorov, Hu, Rohrbach, and
  Singh]{sidorov2020textcaps}
Oleksii Sidorov, Ronghang Hu, Marcus Rohrbach, and Amanpreet Singh.
\newblock Textcaps: a dataset for image captioning with reading comprehension.
\newblock In \emph{Computer Vision--ECCV 2020: 16th European Conference,
  Glasgow, UK, August 23--28, 2020, Proceedings, Part II 16}, pages 742--758.
  Springer, 2020.

\bibitem[Li et~al.(2017)Li, Xiao, Li, Zhou, Yue, and Wang]{li2017person}
Shuang Li, Tong Xiao, Hongsheng Li, Bolei Zhou, Dayu Yue, and Xiaogang Wang.
\newblock Person search with natural language description.
\newblock \emph{arXiv preprint arXiv:1702.05729}, 2017.

\bibitem[Young et~al.(2014)Young, Lai, Hodosh, and Hockenmaier]{young2014image}
Peter Young, Alice Lai, Micah Hodosh, and Julia Hockenmaier.
\newblock From image descriptions to visual denotations: New similarity metrics
  for semantic inference over event descriptions.
\newblock \emph{Transactions of the Association for Computational Linguistics},
  2:\penalty0 67--78, 2014.

\bibitem[Grubinger et~al.(2006)Grubinger, Clough, M{\"u}ller, and
  Deselaers]{grubinger2006iapr}
Michael Grubinger, Paul Clough, Henning M{\"u}ller, and Thomas Deselaers.
\newblock The iapr tc-12 benchmark: A new evaluation resource for visual
  information systems.
\newblock In \emph{International workshop ontoImage}, volume~2, 2006.

\bibitem[Krishna et~al.(2017)Krishna, Zhu, Groth, Johnson, Hata, Kravitz, Chen,
  Kalantidis, Li, Shamma, et~al.]{krishna2017visual}
Ranjay Krishna, Yuke Zhu, Oliver Groth, Justin Johnson, Kenji Hata, Joshua
  Kravitz, Stephanie Chen, Yannis Kalantidis, Li-Jia Li, David~A Shamma, et~al.
\newblock Visual genome: Connecting language and vision using crowdsourced
  dense image annotations.
\newblock \emph{International journal of computer vision}, 123:\penalty0
  32--73, 2017.

\bibitem[Rashtchian et~al.(2010)Rashtchian, Young, Hodosh, and
  Hockenmaier]{rashtchian2010collecting}
Cyrus Rashtchian, Peter Young, Micah Hodosh, and Julia Hockenmaier.
\newblock Collecting image annotations using amazon’s mechanical turk.
\newblock In \emph{Proceedings of the NAACL HLT 2010 workshop on creating
  speech and language data with Amazon’s Mechanical Turk}, pages 139--147,
  2010.

\bibitem[Hudson and Manning(2019)]{hudson2019gqa}
Drew~A Hudson and Christopher~D Manning.
\newblock Gqa: A new dataset for real-world visual reasoning and compositional
  question answering.
\newblock In \emph{Proceedings of the IEEE/CVF conference on computer vision
  and pattern recognition}, pages 6700--6709, 2019.

\bibitem[Zhu et~al.(2016)Zhu, Groth, Bernstein, and Fei-Fei]{zhu2016visual7w}
Yuke Zhu, Oliver Groth, Michael Bernstein, and Li~Fei-Fei.
\newblock Visual7w: Grounded question answering in images.
\newblock In \emph{Proceedings of the IEEE conference on computer vision and
  pattern recognition}, pages 4995--5004, 2016.

\bibitem[Bigham et~al.(2010)Bigham, Jayant, Ji, Little, Miller, Miller, Miller,
  Tatarowicz, White, White, et~al.]{bigham2010vizwiz}
Jeffrey~P Bigham, Chandrika Jayant, Hanjie Ji, Greg Little, Andrew Miller,
  Robert~C Miller, Robin Miller, Aubrey Tatarowicz, Brandyn White, Samual
  White, et~al.
\newblock Vizwiz: nearly real-time answers to visual questions.
\newblock In \emph{Proceedings of the 23nd annual ACM symposium on User
  interface software and technology}, pages 333--342, 2010.

\bibitem[Marino et~al.(2019)Marino, Rastegari, Farhadi, and
  Mottaghi]{marino2019ok}
Kenneth Marino, Mohammad Rastegari, Ali Farhadi, and Roozbeh Mottaghi.
\newblock Ok-vqa: A visual question answering benchmark requiring external
  knowledge.
\newblock In \emph{Proceedings of the IEEE/cvf conference on computer vision
  and pattern recognition}, pages 3195--3204, 2019.

\bibitem[Chen et~al.(2021)Chen, Zhao, Chen, Zhang, Ji, Luo, Xiong, and
  Yu]{chen2021websrc}
Xingyu Chen, Zihan Zhao, Lu~Chen, Danyang Zhang, Jiabao Ji, Ao~Luo, Yuxuan
  Xiong, and Kai Yu.
\newblock Websrc: A dataset for web-based structural reading comprehension.
\newblock \emph{arXiv preprint arXiv:2101.09465}, 2021.

\bibitem[Singh et~al.(2019)Singh, Natarjan, Shah, Jiang, Chen, Parikh, and
  Rohrbach]{singh2019towards}
Amanpreet Singh, Vivek Natarjan, Meet Shah, Yu~Jiang, Xinlei Chen, Devi Parikh,
  and Marcus Rohrbach.
\newblock Towards vqa models that can read.
\newblock In \emph{Proceedings of the IEEE Conference on Computer Vision and
  Pattern Recognition}, pages 8317--8326, 2019.

\bibitem[Biten et~al.(2019)Biten, Tito, Mafla, Gomez, Rusinol, Valveny,
  Jawahar, and Karatzas]{biten2019scene}
Ali~Furkan Biten, Ruben Tito, Andres Mafla, Lluis Gomez, Mar{\c{c}}al Rusinol,
  Ernest Valveny, CV~Jawahar, and Dimosthenis Karatzas.
\newblock Scene text visual question answering.
\newblock In \emph{Proceedings of the IEEE/CVF international conference on
  computer vision}, pages 4291--4301, 2019.

\bibitem[Deng et~al.(2009)Deng, Dong, Socher, Li, Li, and
  Fei-Fei]{deng2009imagenet}
Jia Deng, Wei Dong, Richard Socher, Li-Jia Li, Kai Li, and Li~Fei-Fei.
\newblock Imagenet: A large-scale hierarchical image database.
\newblock In \emph{2009 IEEE conference on computer vision and pattern
  recognition}, pages 248--255. Ieee, 2009.

\bibitem[Xie et~al.(2019)Xie, Lai, Doran, and Kadav]{xie2019visual}
Ning Xie, Farley Lai, Derek Doran, and Asim Kadav.
\newblock Visual entailment: A novel task for fine-grained image understanding.
\newblock \emph{arXiv preprint arXiv:1901.06706}, 2019.

\bibitem[Suhr et~al.(2018)Suhr, Zhou, Zhang, Zhang, Bai, and
  Artzi]{suhr2018corpus}
Alane Suhr, Stephanie Zhou, Ally Zhang, Iris Zhang, Huajun Bai, and Yoav Artzi.
\newblock A corpus for reasoning about natural language grounded in
  photographs.
\newblock \emph{arXiv preprint arXiv:1811.00491}, 2018.

\bibitem[Tan et~al.(2019)Tan, Chan, Aguirre, and Tanaka]{artgan2018}
Wei~Ren Tan, Chee~Seng Chan, Hernan Aguirre, and Kiyoshi Tanaka.
\newblock Improved artgan for conditional synthesis of natural image and
  artwork.
\newblock \emph{IEEE Transactions on Image Processing}, 28\penalty0
  (1):\penalty0 394--409, 2019.
\newblock \doi{10.1109/TIP.2018.2866698}.
\newblock URL \url{https://doi.org/10.1109/TIP.2018.2866698}.

\bibitem[Bulbul et~al.(2018)Bulbul, Cetin, and Dogru]{bulbul2018human}
Erhan Bulbul, Aydin Cetin, and Ibrahim~Alper Dogru.
\newblock Human activity recognition using smartphones.
\newblock In \emph{2018 2nd international symposium on multidisciplinary
  studies and innovative technologies (ismsit)}, pages 1--6. IEEE, 2018.

\bibitem[Kiela et~al.(2020)Kiela, Firooz, Mohan, Goswami, Singh, Ringshia, and
  Testuggine]{kiela2020hateful}
Douwe Kiela, Hamed Firooz, Aravind Mohan, Vedanuj Goswami, Amanpreet Singh,
  Pratik Ringshia, and Davide Testuggine.
\newblock The hateful memes challenge: Detecting hate speech in multimodal
  memes.
\newblock \emph{Advances in Neural Information Processing Systems},
  33:\penalty0 2611--2624, 2020.

\bibitem[Xu et~al.(2017)Xu, Zhao, Xiao, Wu, Zhang, He, and Zhuang]{xu2017video}
Dejing Xu, Zhou Zhao, Jun Xiao, Fei Wu, Hanwang Zhang, Xiangnan He, and Yueting
  Zhuang.
\newblock Video question answering via gradually refined attention over
  appearance and motion.
\newblock In \emph{ACM Multimedia}, 2017.

\bibitem[Pont-Tuset et~al.(2020)Pont-Tuset, Uijlings, Changpinyo, Soricut, and
  Ferrari]{pont2020connecting}
Jordi Pont-Tuset, Jasper Uijlings, Soravit Changpinyo, Radu Soricut, and
  Vittorio Ferrari.
\newblock Connecting vision and language with localized narratives.
\newblock In \emph{Computer Vision--ECCV 2020: 16th European Conference,
  Glasgow, UK, August 23--28, 2020, Proceedings, Part V 16}, pages 647--664.
  Springer, 2020.

\bibitem[Xiao et~al.(2021)Xiao, Shang, Yao, and Chua]{xiao2021next}
Junbin Xiao, Xindi Shang, Angela Yao, and Tat-Seng Chua.
\newblock Next-qa: Next phase of question-answering to explaining temporal
  actions.
\newblock In \emph{Proceedings of the IEEE/CVF Conference on Computer Vision
  and Pattern Recognition}, pages 9777--9786, 2021.

\bibitem[Zhang et~al.(2023{\natexlab{b}})Zhang, Mo, Xu, Si, and
  Kong]{invigdataset}
Hanbo Zhang, Yuchen Mo, Jie Xu, Qingyi Si, and Tao Kong.
\newblock Invig: Interactive visual-language disambiguation with 21k
  human-to-human dialogues.
\newblock \url{https://github.com/ZhangHanbo/invig-dataset},
  2023{\natexlab{b}}.

\bibitem[Nguyen et~al.(2023)Nguyen, Suri, Tsui, Shahules786, Together.xyz, and
  Schuhmann]{laionoigsmall}
Huu Nguyen, Sameer Suri, Ken Tsui, Shahules786, Together.xyz, and Christoph
  Schuhmann.
\newblock The oig small, March 2023.
\newblock URL \url{https://laion.ai/blog/oig-dataset/}.

\bibitem[Honovich et~al.(2022)Honovich, Scialom, Levy, and
  Schick]{honovich2022unnatural}
Or~Honovich, Thomas Scialom, Omer Levy, and Timo Schick.
\newblock Unnatural instructions: Tuning language models with (almost) no human
  labor, 2022.
\newblock URL \url{https://arxiv.org/abs/2212.09689}.

\end{thebibliography}

\appendix

\newpage

\section{Experimental Details}

\subsection{Training Details}
We use the DeepSpeed \cite{rasley2020deepspeed} to accelerate training, and set the BFloat16 as the default model precision.
We report the detailed model training hyperparameters in Table \ref{tab:hyper}.
 
\begin{table}[htbp]
\begin{center}
\resizebox{.7\columnwidth}{!}{
\begin{tabular}{l|ccc}
\toprule
hyperparameters & Pretrain-224 & Pretrain-420 & Finetune \\

\midrule
Env & A100*32 & A100*32 & A100*24 \\
Training steps & 100,000 & 10,000 & 20,000\\
Warmup steps rate & 0.05 & 0.05 & 0.05\\
Warmup lr end & 1e-5 & 1e-6 & 2e-6\\
Optimizer & AdamW & AdamW & AdamW\\
Learning rate & 1e-4 & 1e-5 & 2e-5\\
Learning rate decay & linear & linear & linear\\
Adam $\epsilon$ & 1e-8 & 1e-8 & 1e-8\\
Adam $\beta$ & (0.9, 0.999) & (0.9, 0.999) & (0.9, 0.999)\\
Weight decay & 0.01 & 0.01 & 0.01\\
\bottomrule
\end{tabular}
}
\end{center}
\caption[caption]{Training hyperparameters. Some parameters not use learning rate decay schedule.} 
\label{tab:hyper}
\end{table}

\subsection{Hyper-parameters for Generation}
\label{app:hyper}

During the deployment of all models, we find that for most of them, the performance would be better if we apply a description-first strategy.
That is, before sending the request from the user, by default, we feed a fixed prompt ``Describe the image in detail'' first in the ``0th'' round of the conversation.
After that, the user's instructions will be sequentially processed.
Nevertheless, we found that the quality of generated texts by MiniGPT4 using this description-first strategy is worse than the ones directly generated.
Therefore, for MiniGPT4 \cite{zhu2023minigpt}, we generated the response with its default settings. 
Similarly, for mPLUG-owl \cite{ye2023mplug}, we follow the default parameters presented at \href{http://vlarena.opengvlab.com/}{http://vlarena.opengvlab.com/}.
Detailed settings can be found in \ref{tab:hyper-infer} for different tasks.

\begin{table}[h]
\begin{center}
\resizebox{\linewidth}{!}{
\begin{tabular}{l|ccccccc}
\toprule
& max new tokens & beam size & top-p & top-k & length penalty & no repeat ngram & do sample\\

\midrule
Image Description & 64 & 5 & 1.0  & 1  & -2.0 & 2 & False \\
Open-VQA image & 64 & 5 & 1.0  & 1  & -2.0 & 2 & False \\
Video Description$^*$ & 128 & 1 & 0.9 & 3 & 1.0 & 3 & True \\
Open-VQA video & 128 & 3 & 1.0  & 1  & -1.0 & 3 & False \\
OwlEval Description$^*$ & 128 & 1 & 0.9  & 3  & 1.0 & 3 & True \\
OwlEval & 256 & 3 & 0.9  & 3  & 1.0 & 3 & True \\
demo(ours) & 256 & 3 & 0.9 & 3 & 1.0 & 3 & True \\
\bottomrule
\multicolumn{8}{l}{* The hyperparameters to generate the 0th-round detailed description, if applicable.}
\end{tabular}
}
\end{center}
\caption[caption]{Hyper-parameters for visual question answering evaluation and general-purpose natural language generation with vision inputs respectively. We set hyper-parameters to encourage short response generation for the Open-VQA benchmark. } 
\label{tab:hyper-infer}
\end{table}

\newpage

\section{MME Performance}
\label{ap:mme}

\begin{table*}[h]
\begin{center}
% \small 
\resizebox{\linewidth}{!}{
\begin{tabular}{l|ccccccc}
\toprule
 & BLIP2 \cite{li2023blip}& \specialcell{Instrunct-\\BLIP} \cite{dai2023instructblip}& \specialcell{LLaMA-\\Adapter V2 \cite{gao2023llama}} & \specialcell{mPLUG\\Owl \cite{ye2023mplug}} & MiniGPT4 \cite{zhu2023minigpt}& LLaVA \cite{liu2023llava}& Ours\\
\midrule
Existence & 160.00 & 185.00 & 120.00 & 120.00 & 115.00 & 50.00 & \bf 195.00 \\
Count & 135.00 & 143.33 & 50.00 & 50.00 & 123.33 & 50.00 & \bf 151.67\\
Position & 73.33 & 66.67 & 48.33 & 50.00 & 81.67 & 50.00 & \bf 90.00 \\
Color & 148.33 & 153.33 & 75.00 & 55.00 & 110.00 & 55.00 & \bf 170.00 \\
Poster & \bf 141.84 & 123.81 & 99.66 & 136.05 & 55.78 & 50.00 & 124.83 \\
Celebrity & 105.59 & 101.18 & 86.18 & 100.29 & 65.29 & 48.82 & \bf 118.24 \\
Scene & 145.25 & 153.00 & 148.50 & 135.50 & 95.75 & 50.00 & \bf 164.50 \\
Landmark & 138.00 & 79.75 & 150.25 & 159.25 & 69.00 & 50.00 & \bf 162.00 \\
Artwork & \bf 136.50 & 134.25 & 69.75 & 96.25 & 55.75 & 49.00 & 119.50 \\
OCR & 110.00 & 72.50 & \bf 125.00 & 65.00 & 95.00 & 50.00 & 77.50 \\
\midrule
Perception & 1293.84 & 1212.82 & 972.67 & 967.35 & 866.58 & 502.82 & \bf 1373.23\\
\midrule
Commonsense & 110.00 & \bf 129.29 & 81.43 & 78.57 & 72.14 & 57.14 & 110.71 \\
Numerical & 40.00 & 40.00 & \bf 62.50 & 60.00 & 55.00 & 50.00 & 17.50 \\
Text Translation & 65.00 & 65.00 & 50.00 & \bf 80.00 & 55.00 & 57.50 & 42.50 \\
Code Reasoning & 75.00 & 57.50 & 55.00 & 57.50 & \bf 110.00 & 50.00 & 45.00 \\
\midrule
Cognition & 290.00 & 291.79 & 248.93 & 276.07 & \bf 292.14 & 214.64 & 215.71 \\
\bottomrule
\end{tabular}}
\end{center}
\caption[caption]{Comparison of existing open-sourced multi-modal LLMs on MME benchmark \cite{fu2023mme}. }
\label{tab:vqares}
\end{table*}

\begin{table*}[h]
\begin{center}
% \small 
\resizebox{\linewidth}{!}{
\begin{tabular}{l|ccccccc}
\toprule
 & BLIP2 \cite{li2023blip}& \specialcell{Instrunct-\\BLIP \cite{dai2023instructblip}} & \specialcell{LLaMA-\\Adapter V2 \cite{gao2023llama}} & \specialcell{mPLUG\\Owl \cite{ye2023mplug}} & MiniGPT4 \cite{zhu2023minigpt}& LLaVA \cite{liu2023llava}& Ours\\
\midrule
Existence & 3 & 2 & 4 & 5 & 6 & 7 & 1\\
Count & 3 & 2 & 5 & 6 & 4 & 7 & 1\\
Position & 3 & 4 & 7 & 5 & 2 & 6 & 1\\
Color & 3 & 2 & 5 & 6 & 4 & 7 & 1\\
Poster & 1 & 4 & 5 & 2 & 6 & 7 & 3\\
Celebrity & 2 & 3 & 5 & 4 & 6 & 7 & 1\\
Scene & 4 & 2 & 3 & 5 & 6 & 7 & 1\\
Landmark & 4 & 5 & 3 & 2 & 6 & 7 & 1\\
Artwork & 1 & 2 & 5 & 4 & 6 & 7 & 3\\
OCR & 2 & 5 & 1 & 6 & 3 & 7 & 4\\
\midrule
Perception & 2 & 3 & 4 & 5 & 6 & 7 & 1\\
\midrule
Commonsense & 3 & 1 & 4 & 5 & 6 & 7 & 2\\
Numerical & 5 & 6 & 1 & 2 & 3 & 4 & 7\\
Text Translation & 2 & 3 & 6 & 1 & 5 & 4 & 7\\
Code Reasoning & 2 & 3 & 5 & 4 & 1 & 6 & 7\\
\midrule
Cognition & 3 & 2 & 5 & 4 & 1 & 7 & 6\\
\bottomrule
\end{tabular}}
\end{center}
\caption[caption]{Rankings of existing open-sourced multi-modal LLMs on MME benchmark \cite{fu2023mme}. }
\label{tab:vqares}
\end{table*}

\newpage

\section{Case Study}

\subsection{Image VQA Cases}

\begin{figure}[h]
    \centering
    \includegraphics[width=\textwidth]{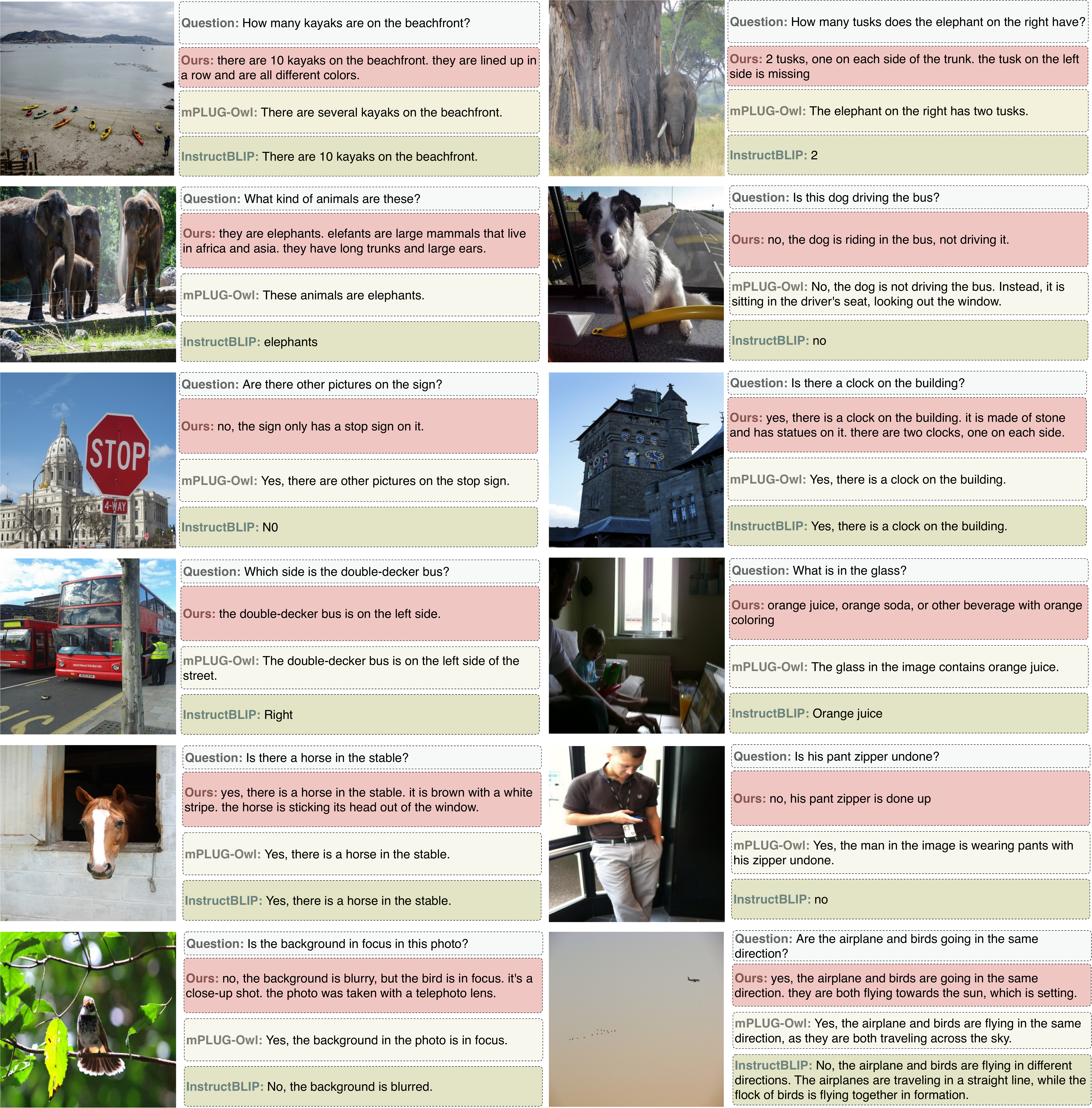}
    \caption{More cases on our Open-VQA image benchmark.}
    \label{fig:more-image-vqa-cases}
\end{figure}

\newpage

\subsection{Video VQA Cases}

\begin{figure}[h]
    \centering
    \includegraphics[width=\textwidth]{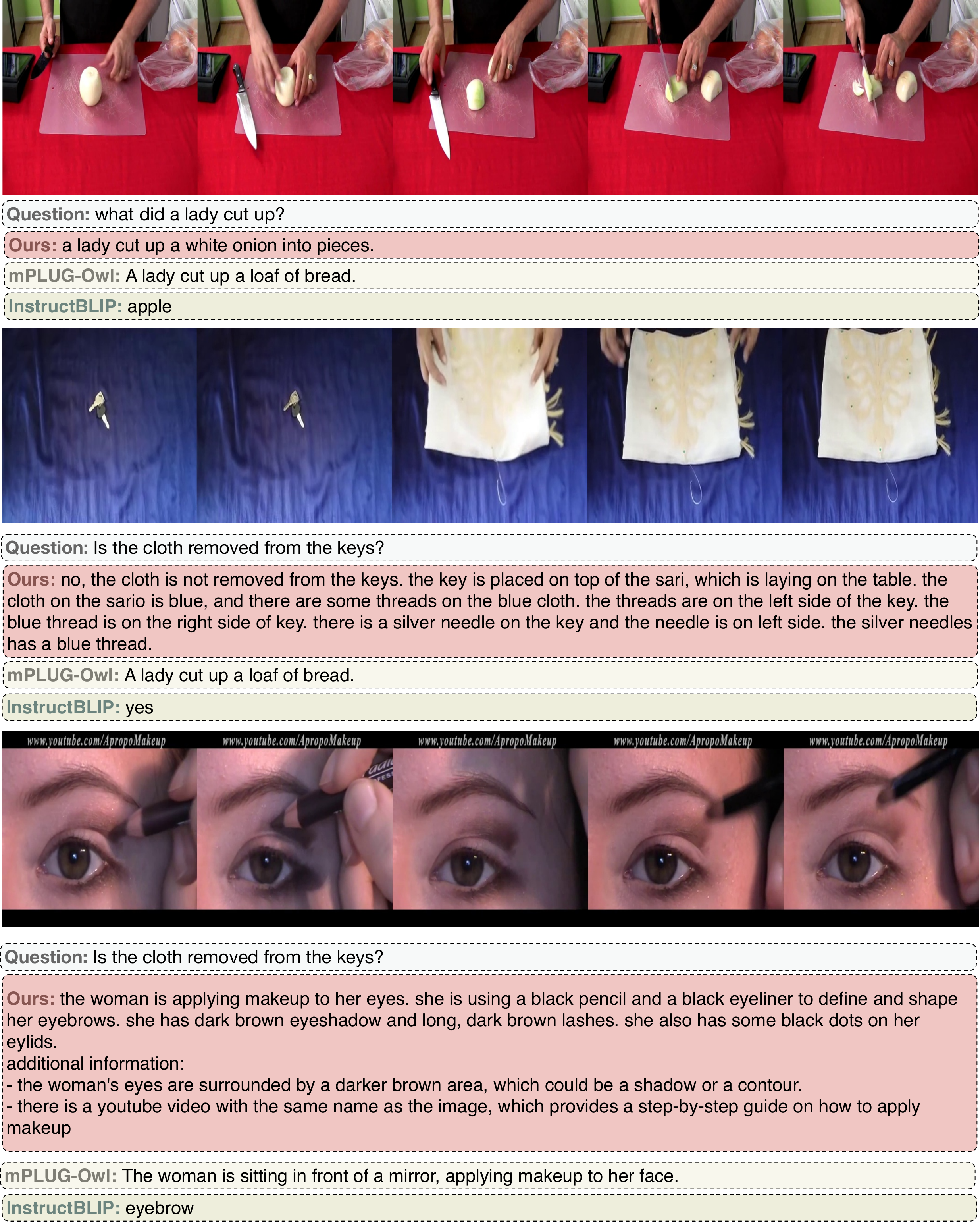}
    \caption{More cases on our Open-VQA video benchmark.}
    \label{fig:more-video-vqa-cases}
\end{figure}

\newpage

\newpage

\section{Training Data}
\label{ap:data}

\subsection{Data \& Tasks}

\begin{table}[h]
    \centering
    \resizebox{.92\linewidth}{!}{
    \begin{tabular}{l|c|c|c|c|c|c}
    \toprule
    Dataset  & Total size & Type & Pretrain & Pretain Ratio & Finetune & Finetune Ratio\\
    \midrule
    BlipCapFilt \cite{li2022blip} & 102.8M & Image-text Pair & \checkmark & 30.525\% & \XSolidBrush& - \\
    CC12M \cite{changpinyo2021conceptual}  & 8.3M & Image-text Pair & \checkmark & 2.465\% & \XSolidBrush& -\\
    CC3M \cite{sharma2018conceptual} & 2.9M  & Image-text Pair & \checkmark & 10.076\% & \XSolidBrush& - \\
    \midrule
    SBU \cite{ordonez2011im2text} & 859.7K & Image Caption& \checkmark & 2.987\% & \XSolidBrush& - \\
    TextCaps \cite{sidorov2020textcaps} & 109.8K & Image Caption  & \checkmark & 0.381\% & \XSolidBrush& - \\
    COCO Caption \cite{chen2015microsoft} & 82.7K & Image Caption & \checkmark & 0.287\% & \XSolidBrush& - \\
    CUHK-PEDES \cite{li2017person} & 34.1K & Image Caption & \checkmark & 0.118\% & \XSolidBrush& -\\
    Flickr30k \cite{young2014image} & 29.8K & Image Caption  & \checkmark & 0.104\% & \XSolidBrush& -\\
    Pexels 110k & 26.2K & Image Caption & \checkmark & 0.091\% & \XSolidBrush& -\\
    LLaVA Caption \cite{liu2023llava} & 23.2K & Image Caption& \XSolidBrush & - & \checkmark & 0.945\%\\
    IAPR TC-12 \cite{grubinger2006iapr} & 20.0K & Image Caption & \checkmark & 0.069\% & \XSolidBrush & -\\
    Visual Genome Caption \cite{krishna2017visual} & 19.6K & Image Caption& \XSolidBrush &- & \checkmark &0.798\%\\
    MiniGPT4 IFT \cite{zhu2023minigpt} & 3.4K & Image Caption& \XSolidBrush &- & \checkmark &0.138\% \\
    Pascal Sentences \cite{rashtchian2010collecting} & 1.0K & Image Caption & \checkmark & 0.003\% & \XSolidBrush &- \\
    \midrule
    VGQA \cite{krishna2017visual} & 1.4M & VQA & \checkmark & 8.711\% & \checkmark & 10.880\% \\
    GQA \cite{hudson2019gqa} & 943.0K & VQA & \checkmark & 5.868\% & \checkmark & 3.999\%  \\
    OCRVQA \cite{mishra2019ocr} & 894.0K & VQA & \checkmark & 5.364\% & \checkmark & 12.349\% \\
    VQAv2 \cite{antol2015vqa} & 443.8K & VQA & \checkmark & 2.761\% & \checkmark & 3.449\% \\
    Visual7W \cite{zhu2016visual7w} & 139.9K & VQA & \checkmark & 0.870\% & \checkmark & 0.593\%\\
    VizWiz \cite{bigham2010vizwiz} & 20.5K & VQA & \checkmark & 0.128\% & \checkmark & 0.087\%\\
    OKVQA \cite{marino2019ok} & 9.0K & VQA & \checkmark & 0.056\% & \checkmark& 0.038\%\\
    TDIUC \cite{kafle2017analysis} & 705.4K & VQA & \checkmark & 4.389\% & \XSolidBrush&- \\
    WebSRC \cite{chen2021websrc} & 131.3K & VQA & \XSolidBrush &- & \checkmark & 1.814\% \\
    LLaVA Reasoning \cite{liu2023llava} & 76.6K & VQA & \XSolidBrush &- & \checkmark & 3.119\%\\
    TextVQA \cite{singh2019towards} & 34.6K & VQA & \XSolidBrush &- & \checkmark & 0.478\%\\
    STVQA \cite{biten2019scene} & 26.0K & VQA & \XSolidBrush &- & \checkmark & 0.359\%\\
    \midrule
    Places365 \cite{zhou2017places} & 1.8M & Classification & \checkmark & 10.921\% & \checkmark& 5.000\%\\
    ImageNet1K \cite{deng2009imagenet} & 1.3M & Classification & \checkmark & 7.887\% & \XSolidBrush &-\\
    SNLI-VE \cite{xie2019visual} & 529.5K & Classification & \checkmark & 3.213\% & \XSolidBrush & -\\
    Visual7W Multi-choice \cite{zhu2016visual7w} & 139.9K & Classification & \checkmark & 0.849\% & \XSolidBrush & -\\
    AirCrowdFood & 100.3K & Classification & \checkmark & 0.609\% & \XSolidBrush & -\\
    NLVR2 \cite{suhr2018corpus} & 86.4K & Classification & \checkmark & 0.518\% & \checkmark & 0.671\%\\
    WikiArt \cite{artgan2018} & 42.5K & Classification & \checkmark & 0.264\% & \checkmark & 0.180\%\\
    HAR \cite{bulbul2018human} &  12.6K & Classification & \checkmark & 0.078\% & \checkmark & 0.053\%\\
    TimeClassification & 11.5K & Classification & \checkmark & 0.072\% & \checkmark & 0.049\% \\
    HatefulMemes \cite{kiela2020hateful} & 8.5K & Classification & \checkmark & 0.026\% & \XSolidBrush & -\\
    \midrule
    MSR-VTT-QA \cite{xu2016msr, xu2017video} & 158.6K & Video VQA& \XSolidBrush &- & \checkmark & 3.137\% \\
    VLN VQA \cite{pont2020connecting} & 31.8K & Video VQA& \XSolidBrush &- & \checkmark & 0.629\% \\   
    NeXT-QA \cite{xiao2021next} & 31.5K & Video VQA& \XSolidBrush &- & \checkmark &  0.623\%\\
    MSVD-QA \cite{chen2011collecting, xu2017video} & 30.9K & Video VQA& \XSolidBrush &- & \checkmark & 0.611\%\\
    \midrule
    SthV2 \cite{goyal2017something} & 168.9K & Video Caption & \XSolidBrush & -& \checkmark & 5.000\% \\
    VLN Caption \cite{pont2020connecting} & 17.6K & Video Caption & \XSolidBrush &- & \checkmark & 5.000\% \\
    \midrule
    LLaVA Instruction \cite{liu2023llava} & 361.4K & Dialog & \XSolidBrush & -& \checkmark & 5.845\%\\
    LLaVA Dialog \cite{liu2023llava} & 256.9K & Dialog & \XSolidBrush & -& \checkmark & 4.155\%\\
    InViG \cite{invigdataset} & 49.9K & Dialog  & \checkmark & 0.310\% & \XSolidBrush &- \\
    \midrule
    Flan V2 \cite{chung2022scaling} &  & Text Instructions& \XSolidBrush &- & \checkmark & 15.000\%\\
    LAION OIG Small \cite{laionoigsmall} & 210.3 & Text Instructions& \XSolidBrush &- & \checkmark &3.884\%\\
    Alpaca GPT4 \cite{wang2022self} & 51.7 & Text Instructions& \XSolidBrush & -& \checkmark & 0.955\%\\
    Unnatural Instruction \cite{honovich2022unnatural} & 8.7 & Text Instructions& \XSolidBrush &- & \checkmark &0.161\%\\
    Baize \cite{xu2023baize} & 601.1 & Text Instructions& \XSolidBrush &- & \checkmark &10.000\%\\
    \bottomrule
    \end{tabular}}
    \vspace{10pt}
    \caption{Training Data.}
    \label{tab:ptdata}
\end{table}

\newpage

\subsection{Prompt Examples}

\begin{table}[ht]
    \centering
    \resizebox{.92\linewidth}{!}{
    \begin{tabular}{l|c|l}
    \toprule
    Dataset  & Type & Prompt Example \\
    \midrule
    BlipCapFilt & Image-text Pair & Describe the image briefly. \\
    CC12M & Image-text Pair & Write a relevant description to pair with the image.\\
    CC3M  & Image-text Pair & Write a relevant description to pair with the image. \\
    \midrule
    SBU & Image Caption & Describe the image.  \\
    TextCaps & Image Caption & Describe the image shortly by reading the texts.\\
    COCO Caption & Image Caption & Describe the image briefly. \\
    CUHK-PEDES & Image Caption & Describe the person in the image.\\
    Flickr30k & Image Caption & Describe the image briefly. \\
    Pexels 110k & Image Caption & Describe the image briefly. \\
    LLaVA Caption & Image Caption & [INSTRUCTION]$^1$ \\
    IAPR TC-12 & Image Caption & Describe the key elements in the image. \\
    Visual Genome Caption &Image Caption & Describe the image in detail.\\
    MiniGPT4 IFT &Image Caption & Describe the image in detail. \\
    Pascal Sentences &Image Caption & Describe the image briefly. \\
    \midrule
    VGQA & VQA & [QUESTION]$^2$ Give a short answer. \\
    GQA & VQA & [QUESTION] Give a short answer.  \\
    OCRVQA & VQA & [QUESTION] Give a short answer.\\
    VQAv2 & VQA & [QUESTION] Give a short answer. \\
    Visual7W & VQA & [QUESTION] Give a short answer.\\
    VizWiz & VQA & [QUESTION] Give a short answer. \\
    OKVQA & VQA & [QUESTION] Give a short answer.\\
    TDIUC & VQA & [QUESTION] Give a short answer. \\
    WebSRC & VQA & Answer the question briefly by reading the webpage. [QUESTION] \\
    LLaVA Reasoning & VQA &  [QUESTION]\\
    TextVQA & VQA & Answer the question shortly by reading the texts. [QUESTION] \\
    STVQA & VQA & [QUESTION] Give a short answer.\\
    \midrule
    Places365 & Classification & Where is this? Answer with a place name.\\
    ImageNet1K & Classification &What is in the image? Answer with its name.\\
    SNLI-VE & Classification &Does the image semantically entail the following text? Text: [HYPO-\\
    & & THESIS]$^3$ Options: 1. neutral 2. entailment 3. contradiction\\
    Visual7W Multi-choice &Classification & Choose the correct answer. Question: [QUESTION] Options: [OP-\\
    & & TIONS]$^4$ \\
    AirCrowdFood & Classification &What food is it?\\
    NLVR2 & Classification &Given the claim "[HYPOTHESIS]", is it True or False?\\
    WikiArt & Classification & What artistic movement or style dose this art picture belong to? Ans-\\
    & & wer with a style name.\\
    HAR & Classification & What is the person doing? Answer shortly.\\
    TimeClassification & Classification &What is the time now? Give a short answer.\\
    HatefulMemes & Classification &Is "[MEME]$^5$" a hateful meme? Answer with Yes or No.\\
    \midrule
    MSR-VTT-QA &Video VQA&[QUESTION] Give a short answer. \\
    VLN VQA &Video VQA&[QUESTION] Give a short answer.\\   
    NeXT-QA &Video VQA& [QUESTION] Give a short answer.\\
    MSVD-QA &Video VQA& [QUESTION] Give a short answer. \\
    \midrule
    SthV2 &Video Caption&Describe the video shortly.\\
    VLN Caption &Video Caption&Describe the video in detail.\\
    \midrule
    LLaVA Instruction & Dialog & [DIALOG]$^6$ \\
    LLaVA Dialog & Dialog &[DIALOG]\\
    InViG & Dialog & [DIALOG]\\
    \midrule
    Flan V2 & Text Instructions & [INSTRUCTION]\\
    LAION OIG Small &Text Instructions &[INSTRUCTION]\\
    Alpaca GPT4 &Text Instructions & [INSTRUCTION]\\
    Unnatural Instruction &Text Instructions &[INSTRUCTION]\\
    Baize & Text Instructions &[INSTRUCTION]\\
    \bottomrule
    \multicolumn{3}{l}{\small 1. We directly use the original instructions if it is applicable.}\\
    \multicolumn{3}{l}{\small 2. [QUESTION] means the original question from the dataset. We do some post process, e.g., add the ``?'' if it is missing.}\\
    \multicolumn{3}{l}{\small 3. [HYPOTHESIS] means the statement in the SN-LIVE dataset.}\\
    \multicolumn{3}{l}{\small 4. [OPTIONS] means the options list provided by the dataset.}\\
    \multicolumn{3}{l}{\small 5. [MEME] is the meme included in the Hateful Memes dataset.}\\
    \multicolumn{3}{l}{\small 6. We split each dialog into rounds, and prompt the model to speak for the next round based on the dialog history.}\\
    \end{tabular}}
    \vspace{10pt}
    \caption{Prompt examples of different datasets.}
    \label{tab:prompts}
\end{table}

\newpage

\section{OwlEval Cases}
\label{app:ablation_case}

\begin{figure}[h]
    \centering
    \includegraphics[width=.85\textwidth]{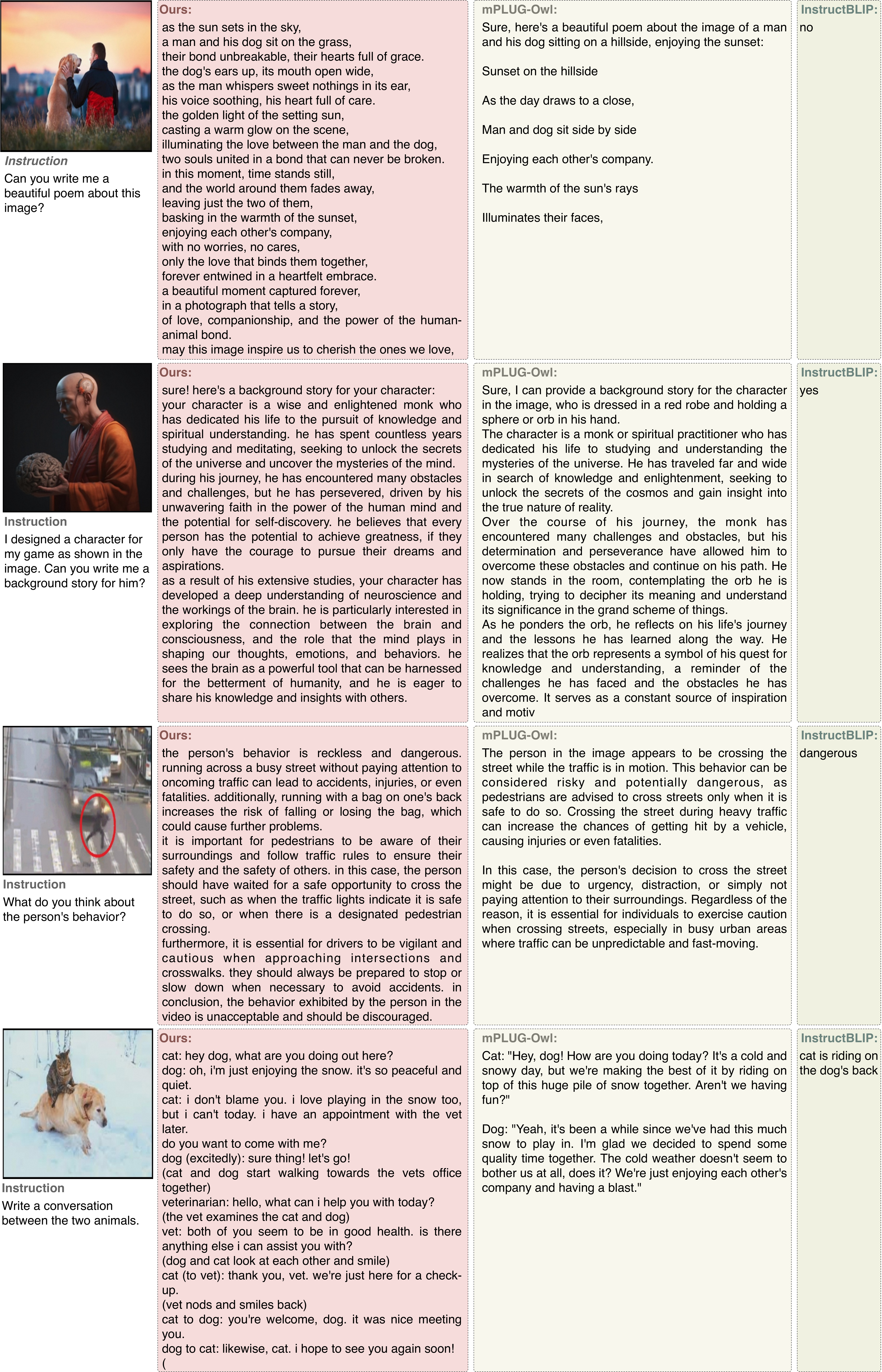}
    \caption{More cases on our OwlEval benchmark.}
    \label{fig:more-owl-cases2}
\end{figure}

\newpage\phantom{blabla}

\begin{figure}[h]
    \centering
    \includegraphics[width=.85\textwidth]{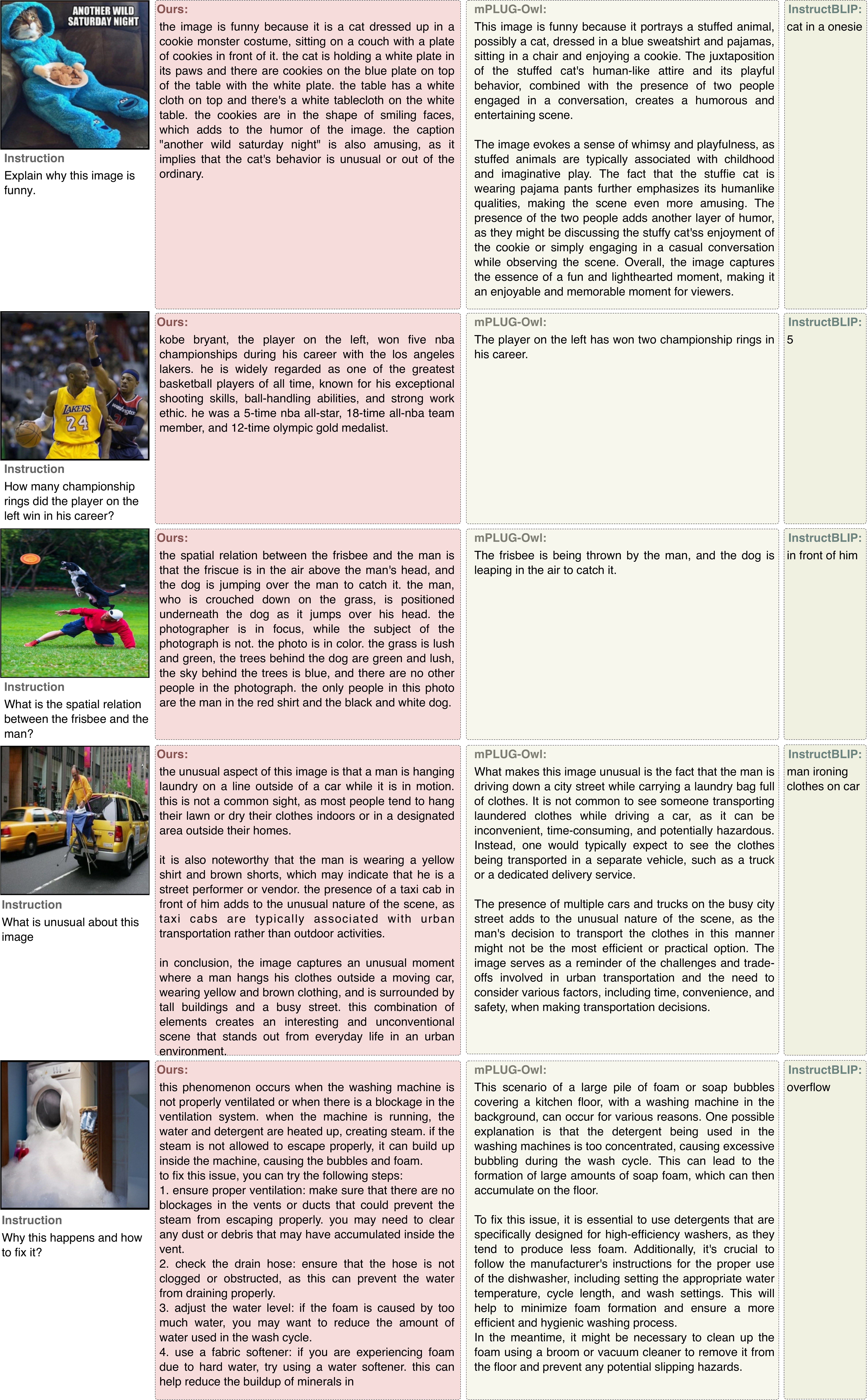}
    \caption{More cases on our OwlEval benchmark.}
    \label{fig:more-owl-cases1}
\end{figure}

% TODO

%%%%%% showcase multi-turn dialogs

\newpage

\section{Open Demonstrations}
\label{sec:open-demo}

\subsection{Multi-turn Dialog}

\begin{figure}[h]
    \centering
    \includegraphics[width=0.95\textwidth]{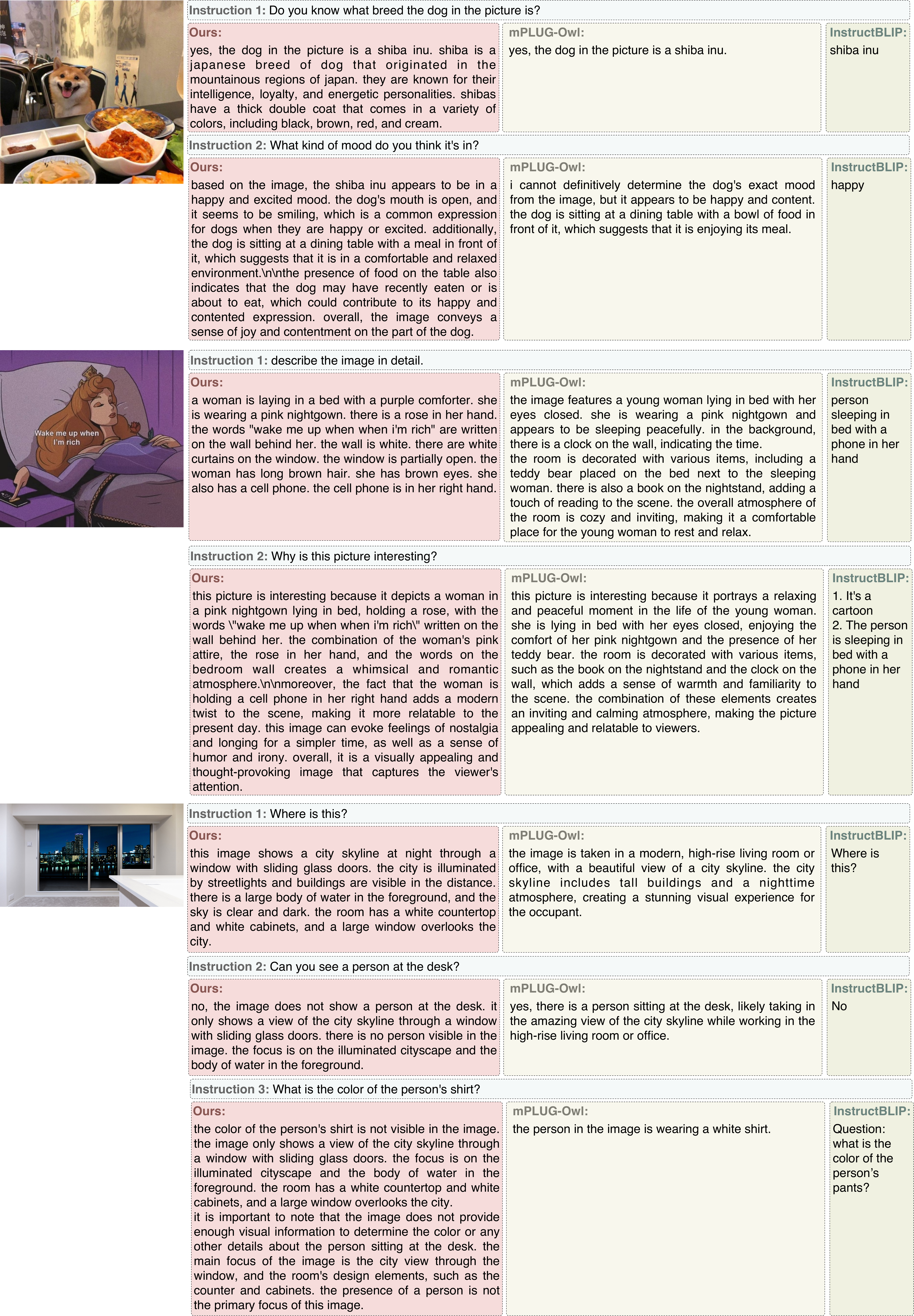}
    \caption{Multi-turn dialog demonstrations.}
    \label{fig:multi-turn-demos}
\end{figure}

\newpage

\subsection{Multi-lingual Response}

\begin{figure}[h]
    \centering
    \includegraphics[width=0.85\textwidth]{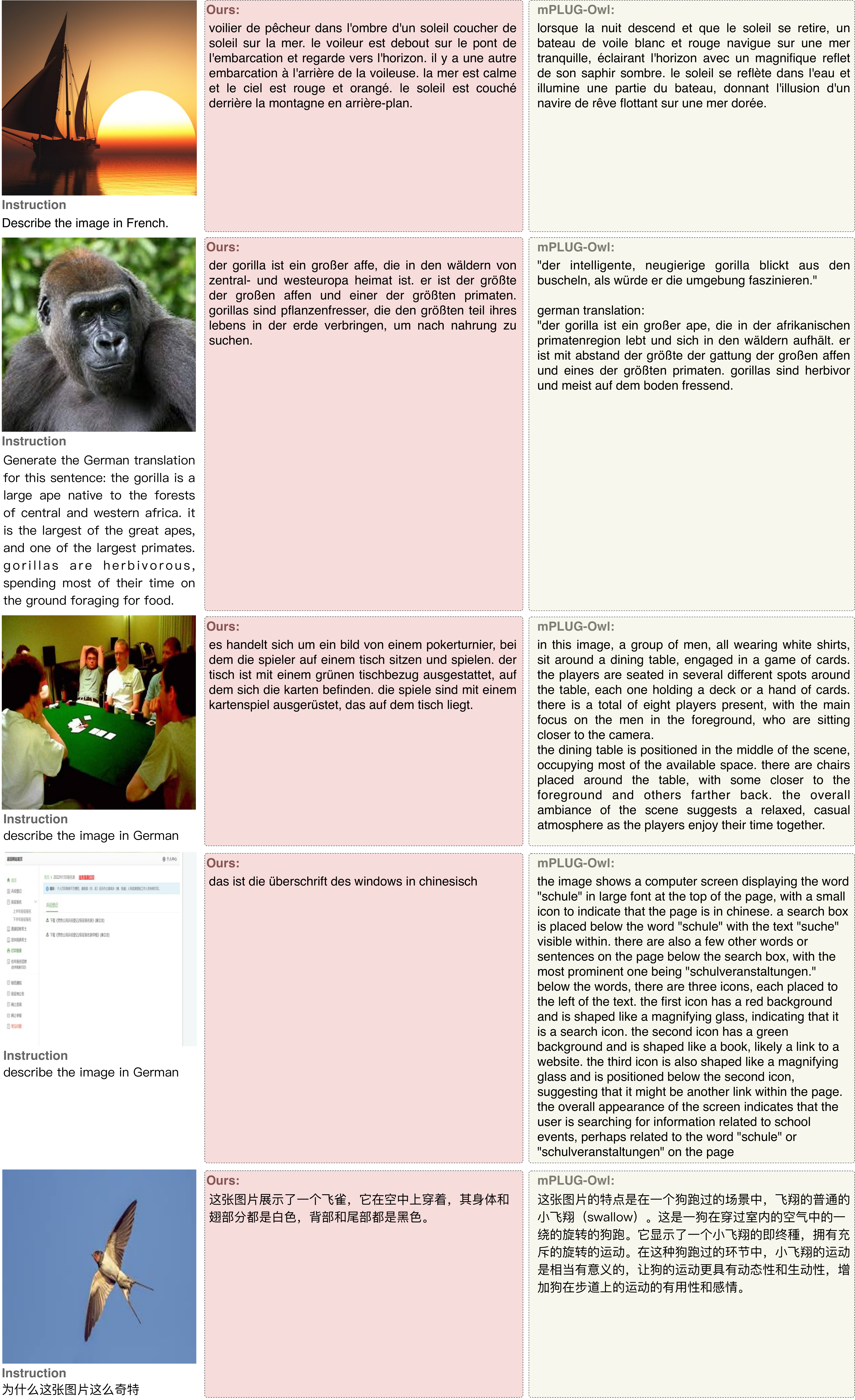}
    \caption{Multi-lingual demonstrations.}
    \label{fig:multi-lingual-demos}
\end{figure}

\newpage

\subsection{Instruction-following Ability}

We also demonstrate the instruction-following ability of different models.
We can see that both \ours and mPLUG-owl can follow instructions correctly to some extent.
Yet, InstructBLIP is not sensitive to different instructions.

\begin{figure}[h]
    \centering
    \includegraphics[width=\textwidth]{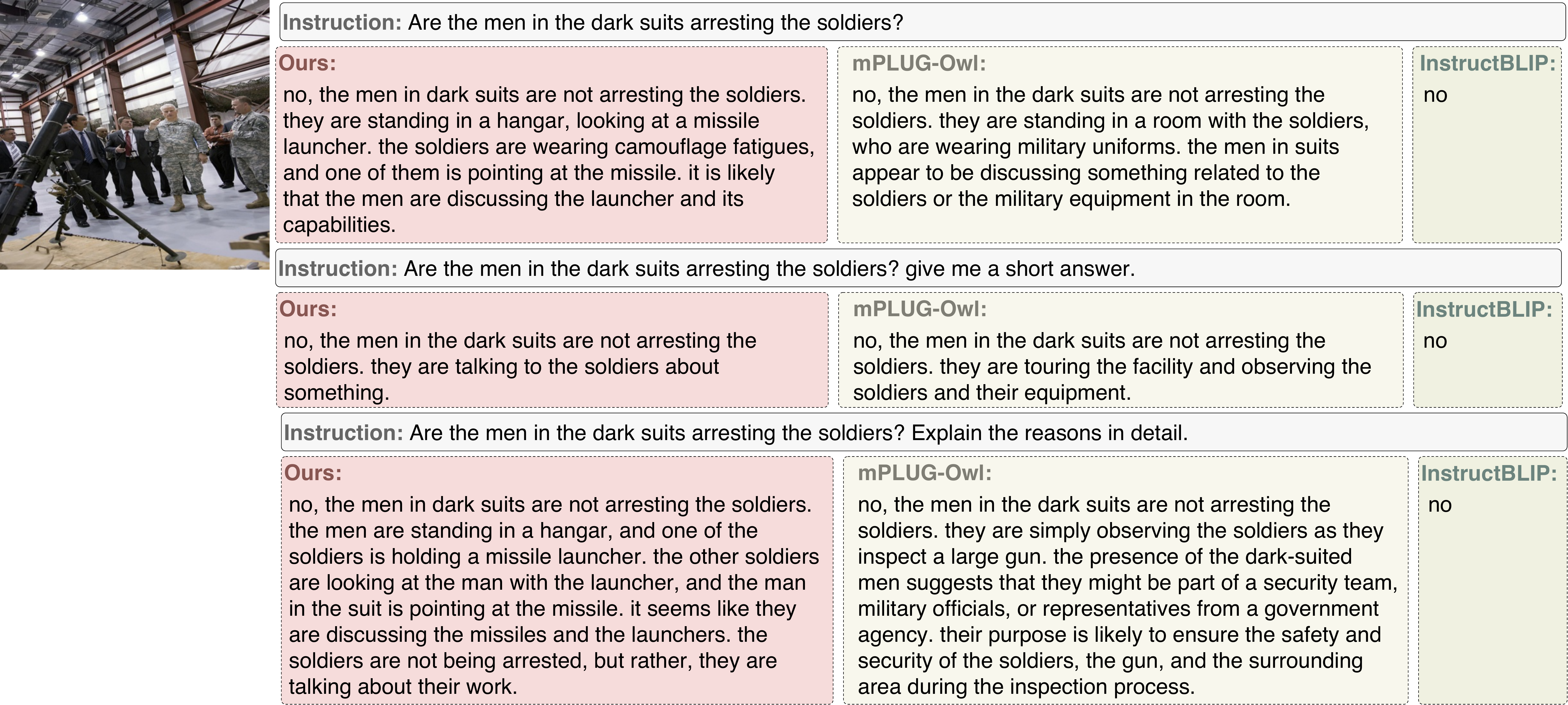}
    \caption{Demonstration of instruction-following ability.}
    \label{fig:multi-lingual-demos}
\end{figure}

\end{document}